\documentclass[letterpaper,journal]{IEEEtran}
\usepackage{amsmath,amsfonts}
\usepackage{algorithm}
\usepackage{algorithmic}
\usepackage{array}
\usepackage[caption=false,font=normalsize,labelfont=sf,textfont=sf]{subfig}
\usepackage{textcomp}
\usepackage{stfloats}
\usepackage{url}
\usepackage{verbatim}
\usepackage{graphicx}
\usepackage{cite}

\usepackage{booktabs}  
\usepackage{multirow}  
\usepackage{lineno} 
\usepackage{colortbl} 
\usepackage{adjustbox} 
\usepackage{bm}
\usepackage{amssymb}
\usepackage{hyperref} 
\usepackage[T1]{fontenc}
\usepackage{threeparttable}    

\usepackage{soul, color, xcolor}

\begin{document}

\title{EvoDR: Evolving Dispatching Rules via Large Language Model for Dynamic Flexible Assembly Flow Shop Scheduling}

\author{Junhao Qiu, Haoyang Zhuang, Fei Liu, Jianjun Liu, Qingfu Zhang,~\IEEEmembership{Fellow,~IEEE}
        
\thanks{This work was supported by the Research Grants Council of the Hong Kong SAR, China (CityU11217325) and the Natural Science Foundation of China (62276223, 52375489) (Corresponding author: Qingfu Zhang).}
\thanks{Junhao Qiu, Fei Liu, and Qingfu Zhang are with the Department of Computer Science, City University of Hong Kong, Hong Kong, China (email: junhaoqiu2-c@my.cityu.edu.hk; fliu36-c@my.cityu.edu.hk; qingfu.zhang@cityu.edu.hk).}
\thanks{Haoyang Zhuang and Jianjun Liu are with the School of Electromechanical Engineering, Guangdong University of Technology, Guangzhou, China (email: zhuanghaoyang@mails.gdut.edu.cn; jianjun.liu@gdut.edu.cn).}
\thanks{\copyright~2026 IEEE. Accepted for publication in IEEE Transactions on Evolutionary Computation. DOI: 10.1109/TEVC.2026.3751778.}
}

\markboth{Journal of \LaTeX\ Class Files,~Vol.~14, No.~10, JUNE~2026}%
{Shell \MakeLowercase{\textit{et al.}}: A Sample Article Using IEEEtran.cls for IEEE Journals}


\maketitle

\begin{abstract}
Dynamic flexible assembly flow shop scheduling with multi-product delivery, often with kitting supply and machine-level flexibility, can be found in many applications. It is difficult to design rules manually for it. 
Genetic programming (GP) has been used to automatically generate dispatching rules, enabling responsive scheduling while meeting high responsiveness demands. However, GP methods are dependent on fixed terminal sets and have poor interpretability. In this work, we develop an evolving dispatching rules framework (EvoDR) that leverages the semantic understanding and generation capabilities of large language models to achieve cross-domain integration of rule design and scheduling knowledge. In our proposed framework, multi-stage assembly supply decisions are modeled as priority sorting of directed edges based on heterogeneous graphs. A dual-expert co-evolution mechanism is proposed, which allows LLM-A to generate code and LLM-S conduct scheduling analysis and reflection. Guided by improvements in hybrid evaluation, adaptive rules that fit dynamic features are continuously evolved. Experimental results show that EvoDR can achieve lower average tardiness than other automatically designed state-of-the-art approaches. It also outperforms expert-designed competitors across 480 instances over 24 distinct scenarios with varying resource configurations and disturbance levels, and EvoDR demonstrates robust adaptability across diverse operating conditions.

\end{abstract}


\section{Introduction}
\label{sec:introduction}
\IEEEPARstart{V}{olatile} order patterns and increasingly complex kitting demands challenge the coordination and production capacity of modern manufacturing systems~\cite{qin2021self, zhang2022distributed}. The flexible assembly flow shop scheduling problem (FAFSP) models the supply relationship between part processing and product assembly~\cite{liu2025recent,komaki2019flow} and is widely applicable in industries such as home appliance manufacturing~\cite{ elyasi2024imperialist}, aerospace~\cite{wu2018two}, machinery manufacturing~\cite{kong2024energy}, and plastics~\cite{allahverdi2015two}. 
Under the constraints of a hierarchical product structure, the demand for multi-product delivery necessitates centralized dispatching of different product families. 
Moreover, the combination of flexible processing capabilities and structure-delivery dual kitting constraints further complicates the operational supply network~\cite{qiu2024novel}, thus requiring coordinated decision-making across multiple stages.

For the traditional static FAFSP, where process information is fully known in advance, researchers have proposed various solution methods. These include exact algorithms such as mathematical programming~\cite{da2024production} and branch and bound~\cite{fattahi2014branch, komaki2019flow}. 
However, these methods rely on the ideal assumption of a stable production environment and require substantial offline computation~\cite{lei2023large}. 
Consequently, these methods cannot adapt to real time uncertainties such as dynamic order arrivals, making them inadequate for dynamic scheduling scenarios that require immediate responsiveness.

In dynamic environments, disturbances often require the rescheduling of job priorities in the cache, directly affecting previously established strict supply relationships~\cite{branke2005anticipation}.
Existing dynamic scheduling methods for manufacturing systems are primarily categorized into proactive and reactive approaches~\cite{ouelhadj2009survey}. 
Proactive scheduling, also known as robust scheduling, typically constructs initial solutions with a certain degree of disturbance resistance by inserting idle time, though often at the cost of scheduling performance~\cite{yan2024learning, lu2012robust}.
Reactive scheduling offers greater flexibility and adaptability, adjusting through real-time decision-making in response to each dynamic event. 
This real-time adjustment is typically achieved by applying priority dispatching rules (PDRs)~\cite{qiu2024novel, yang2008predictive}. These rules are favored for their high computational efficiency and rapid response, making them well-suited for real-time decision-making in dynamic environments~\cite{meng2026Niching}.
However, as heuristic methods, the design of PDRs is resource intensive, relying on substantial domain expertise and iterative trial-and-error validation.
\IEEEpubidadjcol

Automatically designing PDRs given the problem classes is an effective approach to reduce manual reliance~\cite{zhang2023survey, burke2013hyper}. Genetic programming (GP) is one of the most well-known methods in the field of heuristic automatic design for scheduling problems~\cite{zhang2023survey, nguyen2017genetic}. Early work by~\cite{tay2008evolving} used GP to design multi-objective PDRs aimed at optimizing makespan, tardiness, and flowtime for the flexible job shop scheduling problem (FJSP). In dynamic FJSP,~\cite{zhang2020evolving} integrated feature selection into a GP-based hyperheuristic framework, successfully designing more interpretable scheduling heuristics with fewer unique features and reduced complexity.~\cite{ozturk2019extracting} considered the inclusion of scheduling environment information when constructing the terminal set in genetic expression programming (GEP). 
However, these methods typically rely on a predefined set of terminal symbols or mutation actions to construct and evolve heuristics, which necessitates prior specification of the search space and may implicitly incorporate problem specific knowledge into the algorithm design process~\cite{o2010open}. 

Recently, applying large language models (LLMs) for automated heuristic design (AHD) has achieved notable success in various fields~\cite{liu2026systematic}, including combinatorial optimization~\cite{EOH,meoh,li2025llm,reevo,xu2025autoep}, mathematics~\cite{funsearch}, and machine learning~\cite{mo2025autosgnn}. A common paradigm is to integrate LLMs as heuristic designers within evolutionary search, neighborhood search, and other iterative search frameworks~\cite{qiu2026e2oc}.
Notable examples include EoH~\cite{EOH} which evolves both the reasoning process and the code for effective automation. These studies highlight the significant potential of such methods in flow shop and job shop scheduling, demonstrating their broad future applicability.

LLMs have demonstrated strong capabilities in code generation and logical reasoning, yet significant challenges remain in automatically designing dynamic PDRs capable of understanding complex constraints and perturbations~\cite{yu2026automated}: (1) Semantic perception of uncertainty and rules in dynamic scheduling. LLM-generated PDRs react in real time to fluctuating order demands while accounting for factors such as order priority, delivery deadlines, and resource availability. Achieving accurate semantic profiling for rules and dynamic manufacturing system states is of paramount importance to guarantee the robustness of LLM-generated PDRs~\cite{zhang2020evolving}.
(2) Asynchronous integrated scheduling with multi-level PDRs in hybrid processing–assembly systems.
In practical production environments, processing and assembly decisions occur asynchronously, creating decision misalignment. Processing aims to reduce changeover losses and improve efficiency to ensure complete product sets, thereby preventing assembly idle time~\cite{da2024production}. In contrast, assembly prioritizes coordinating different products within the same order to enable consolidated delivery and avoid order delays~\cite{qiu2024multi}.
Therefore, the ability to automatically generate asynchronous, integrated PDRs that incorporate heterogeneous decision preferences and dynamic system semantics is essential for balancing multi-product delivery performance and resource utilization. 

The contributions of this paper are summarized as below.

\begin{enumerate}
    \item  We formulate a heterogeneous graph-based Markov decision process (MDP) for the dual kitting constrained FAFSP, which jointly represents assembly supply and multi-product delivery decisions as directed edges to resolve multi-stage hierarchical conflicts. Furthermore, a feature fitting rule evolution mechanism is designed to generate PDRs that dynamically adapt to system features, enhancing the generalization ability of perturbation scenarios.

    \item We propose an LLM-assisted evolving dispatching rule framework (EvoDR) for dynamic FAFSP. It integrates a dual-expert mechanism, where LLM-A generates rules and LLM-S conducts scheduling analysis and semantic profiling, enabling cross-domain fusion of algorithmic and scheduling knowledge. 
    Furthermore, an elite knowledge guided initialization mechanism injects high-level design knowledge to produce high-quality initial PDRs. We also introduce a hybrid evaluation strategy, combining objective indicators with empirical evaluation to accurately assess PDR's adaptability in dynamic multi-level manufacturing systems, supporting targeted performance improvement.

    \item We conduct a series of experiments on industrial and generated instances to comprehensively evaluate EvoDR. The performance and cost of EvoDR are analyzed in experiments with different LLMs and temperature settings. 
    Then, we demonstrate that EvoDR outperforms many existing AHD methods. 
    Moreover, it surpasses widely used manually designed PDRs for dynamic FAFSP across 480 instances in 24 different scenarios defined by varying resource load levels, disturbance intensities, and order conditions.

\end{enumerate}

The remainder of this paper is organized as follows. 
Section~\ref{sec:Problem} introduces the problem formulation. Section~\ref{sec:method} presents the details of EvoDR. Section~\ref{Exp:experience} reports a series of computational experiments and discusses the corresponding results. Section~\ref{conclusion} concludes the paper, while the Appendix provides the related-work review, the detailed encoding and instance-generation specifications, the optimization formulation, and supplementary analyses of the experiments.
\section{Preliminary}
\label{sec:Problem}

\subsection{Problem Description and Mathematical Model}
The dynamic FAFSP under multi-product delivery requirements represents a shift from optimizing isolated product chains to managing a sophisticated, real-time supply network. 
Unlike traditional FAFSP, where completion depends on a fixed processing-assembly sequence, it necessitates that multiple heterogeneous products within a single order be synchronized for collective delivery. 
These hierarchical dual kitting constraints introduce the critical challenge of kit-completeness, where an order's tardiness is strictly determined by its latest finished component, causing any local disturbance or machine-level delay to propagate across the entire order~\cite{qiu2024multi}. 
Furthermore, the inherent heterogeneity in process routes, machine eligibility, and quantity demands across different products creates substantial load imbalances and divergent completion times. The complexity is further amplified in dynamic scenarios by unpredictable order arrivals, which necessitate the simultaneous coordination of product-level kitting constraints and order-level synchronization requirements. Consequently, the scheduling objective centers on balancing real-time resource utilization with the stringent demands of multi-product synchronization to minimize total order tardiness within a highly coupled and volatile environment.

The notation of the mathematical model is illustrated in Table \ref{Notations_model}. The objective, as defined in Eq.~\ref{eqobj}, is to minimize the total tardiness in multi-product delivery orders.

\begin{table}[h]
\caption{Notations of mathematical model.}\label{Notations_model}
\centering
\resizebox{\columnwidth}{!}{
\begin{tabular}{ll}
\toprule
\multicolumn{2}{l}{\textbf{Indices}}                                                                                                                         \\
$i$           & Order index, $i\in I$                                                                         \\
$p$           & Product index, $p\in P$                                                                         \\
$j$           & Job index, $j_1, j_2\in J$                                                                        \\
$m$           & Machine index, $m\in  M $                                                                                   \\
\multicolumn{2}{l}{\textbf{Parameters}}                                                                                                                      \\
${{Q}_{j, m}}$    & 1 if the machine $m$ is qualified to process the job $j$, 0 otherwise                                                           \\
${{P}_{j_1,j_2}}$   & 1 if job $j_1$ is the immediate predecessor of job $j_2$, 0 otherwise                                                              \\
${{PT}_{j, m}}$    & Processing time of job $j$ on machine $m$                 \\
${{A}_{p, j}}$    & 1 if job $j$ is an assembly task for the product $p$, 0 otherwise                                                          \\
${{O}_{i, p}}$   & 1 if product $p$ belongs to order $i$, 0 otherwise.                                                                                 \\
${{DT}_{i}}$      & Delivery time of order $i$                                                                                                      \\
${{AT}_{i}}$      & Arrival time of order $i$                                                                                                       \\
${{ST}_{j_1,j_2}}$  & Setup time for job $j_1$ switch to job $j_2$                                                                                    \\
$V$    & A large number                                                                      \\
\multicolumn{2}{l}{\textbf{Decision variables}}                                                                                                                       \\
${{ft}_{i}}$      & Completion time of order $i$                                                                                                     \\
${{ct}_{j}}$      & Completion time of job $j$                                                                                                     \\
${{st}_{j}}$      & Start time of processing for job $j$                                                                                             \\
${{x}_{j_1,j_2,m}}$ &  Binary variable, equal to 1 if job $j_1$ is processed before job $j_2$ \\
  & on machine $m$, and 0 otherwise
\\ 
\bottomrule     
\end{tabular}}
\end{table}

\label{formulation}
\begin{equation}
\label{eqobj}
\min \sum_{i \in I} \max \left( 0, ft_{i} - DT_{i} \right)
\end{equation}
Subject to:
\begin{equation}
\label{constant1}
{x}_{j_1,j_2,m}\le {{Q}_{j_2,m}},\forall j_1\in \{0\}\cup J,j_2 \in J,m\in M
\end{equation}
\begin{equation}
\label{constant2}
\sum_{j \in J}{{{x}_{0,j,m}}}=1,\forall m\in M
\end{equation}
\begin{equation}
\label{constant3}
\sum_{m \in M}{\sum_{j_1 \in \{0\}\cup J}{{{x}_{{{j}_{1}},{{j}_{2}},m}}=1,\forall j_2 \in J}}
\end{equation}
\begin{equation}
\label{constant4}
\sum_{j_1 \in \{0\}\cup J}{{{x}_{{{j}_{1}},{{j}_{2}},m}}}-\sum_{j_1 \in \{0\}\cup J}{{{x}_{{{j}_{2}},{{j}_{1}},m}}}=0,\forall j_2 \in J,m\in M
\end{equation}
\begin{equation}
\label{constant5}
\begin{aligned}
st_{j_2}\le ct_{j_2}-PT_{j_2,m}+V\times \left( 1-\sum_{j_1 \in \{0\}\cup J}{x_{j_1,j_2,m}} \right), \\
\forall j_2 \in J,m \in M
\end{aligned}
\end{equation}
\begin{equation}
\label{constant6}
\begin{aligned}
ct_{j_2}\ge  ct_{j_1} + PT_{j_2,m}+ST_{j_1,j_2}+V\times \left( x_{j_1,j_2,m} -1 \right),\\
\forall j_1 \in \{0\}\cup J, j_2 \in J,m \in M
\end{aligned}
\end{equation}
\begin{equation}
\label{constant7}
\begin{aligned}
ct_{j_2}\ge \sum_{m \in M}{\sum_{j_3 \in \{0\}\cup J}{\left( x_{j_3,j_2,m} \times \left( PT_{j_2,m} + ST_{j_3,j_2} \right) \right)}}\\
+ P_{j_1,j_2}\times ct_{j_1},
\forall j_1,j_2 \in J
\end{aligned}
\end{equation}
\begin{equation}
\label{constant8}
\begin{aligned}
ft_i \ge ct_j +V \times \left( A_{p,j} + O_{i,p}-2 \right), \\
\forall i \in I,p\in P,j \in J
\end{aligned}
\end{equation}
\begin{equation}
\label{constant9}
\begin{aligned}
st_j\ge \left( A_{p,j} + O_{i,p}-2 \right)\times V+AT_i, \\
\forall i \in I,p\in P,j \in J
\end{aligned}
\end{equation}

Constraints \ref{constant1}--\ref{constant4} define machine routing and flow conservation, where $j=0$ denotes a virtual start job with $ct_0=0$. Constraints \ref{constant5} and \ref{constant6} enforce temporal continuity and eliminate scheduling conflicts on individual machines, while constraint \ref{constant7} captures the hierarchical coupling to satisfy process precedence routes. 
For order-level coordination, Constraint \ref{constant8} models the multi-product delivery synchronization, where an entire order's completion time is determined by its latest finished product. 
Constraint \ref{constant9} introduces the dynamic availability constraint, restricting the processing of any constituent job from starting before its corresponding order arrives at the workshop. 
Finally, all timeline and delay-related decision variables ($ft_i$, $ct_j$, and $st_j$) are restricted to non-negative real values, as time cannot flow backward.

\subsection{Markov Decision Process}
\label{sec:mdp}
The dynamic FAFSP is formulated as an event-driven MDP where decisions are triggered at $t=0$ or upon job completion. At each step, a PDR filters the feasible action set to execute integrated processing and assembly operations until the schedule is complete. Job scheduling and resource assignment are jointly modeled as the selection of directed $j$-$m$ edges in a heterogeneous graph.

\textit{State $s_t$:} The system state $s_t \in \mathbb{R}^9$ (detailed in Table~\ref{state_deatures}) captures the dynamic state features across orders, job-machine arcs, and machines. These features are recomputed at each step $t$ to reflect order arrivals and resource fluctuations.

\textit{Actions $a_t$:} An action $a_t \in A_t$ refers to the set of feasible directed edges $(j,m)$ at the current decision time that satisfy the constraints. A pair is feasible only if job $j$ is released (predecessors finished) and machine $m$ is idle. The action space is dynamic, expanding with new order arrivals and shrinking as assignments are made.

\textit{Transition:} Transitions occur as the selected PDR iteratively assigns pairs until the feasible set is exhausted. The system then advances to the next decision step defined by the earliest job completion or new order arrival.

\begin{table}[t]
\caption{State features at decision step $t$.}\label{state_deatures}
\centering
\resizebox{1\columnwidth}{!}{
\begin{tabular}{lcl}
\toprule
\textbf{Type} & \textbf{Index} & \textbf{Description}                                   \\
\midrule
Order
& $\bm{f}_{1}$           & Processing status binary variables of each job $j\in J_t$ at step $t$,         \\
 &             &   1 if job is processed or not, 0 otherwise.                 \\
& $\bm{f}_{2}$           &  Actual processing time at step $t$, cumulative  time if the job has            \\
&             &  been processed, otherwise average $PT_{i}$ on available machines. \\
& $\bm{f}_{3}$           &  Number of available processing machines for job $j$ at step $t$.           \\ 
\hline
Arc
& $\bm{f}_{4}$           & Arrival time for each job $j\in J_t$.       \\
& $\bm{f}_{5}$           &  Delivery time for each job $j\in J_t$. \\
& $\bm{f}_{6}$           &  Processing time for each job $j\in J_t$.            \\  
\hline
Machine
& $\bm{f}_{7}$           & Next available time for each machine.       \\
& $\bm{f}_{8}$           &  Number of qualified processing jobs for each machine. \\
& $\bm{f}_{9}$           &  Utilization of each machine.           

\\ \bottomrule    
\end{tabular}}
\end{table}

Based on this decision process, the quality of a dispatching rule is measured by the tardiness it induces on the training instances. Let $\mathcal{X}$ be the set of training instances, where each $\mathrm{x}\in\mathcal{X}$ specifies a manufacturing configuration. The tardiness of a rule $\pi$ on instance $\mathrm{x}$ is denoted by $F(\mathrm{x}\mid\pi)$, and its average over the instance set by $F(\mathcal{X}\mid\pi)$. The optimization objective is to find $\pi^\ast=\arg\min_{\pi\in\mathbf{\Pi}}F(\mathcal{X}\mid\pi)$, where $\mathbf{\Pi}$ is the feasible PDR space.

\section{Proposed EvoDR}
\label{sec:method}

\begin{figure*}[t]
\centering
\makebox[\textwidth][c]{\includegraphics[width=.95\textwidth]{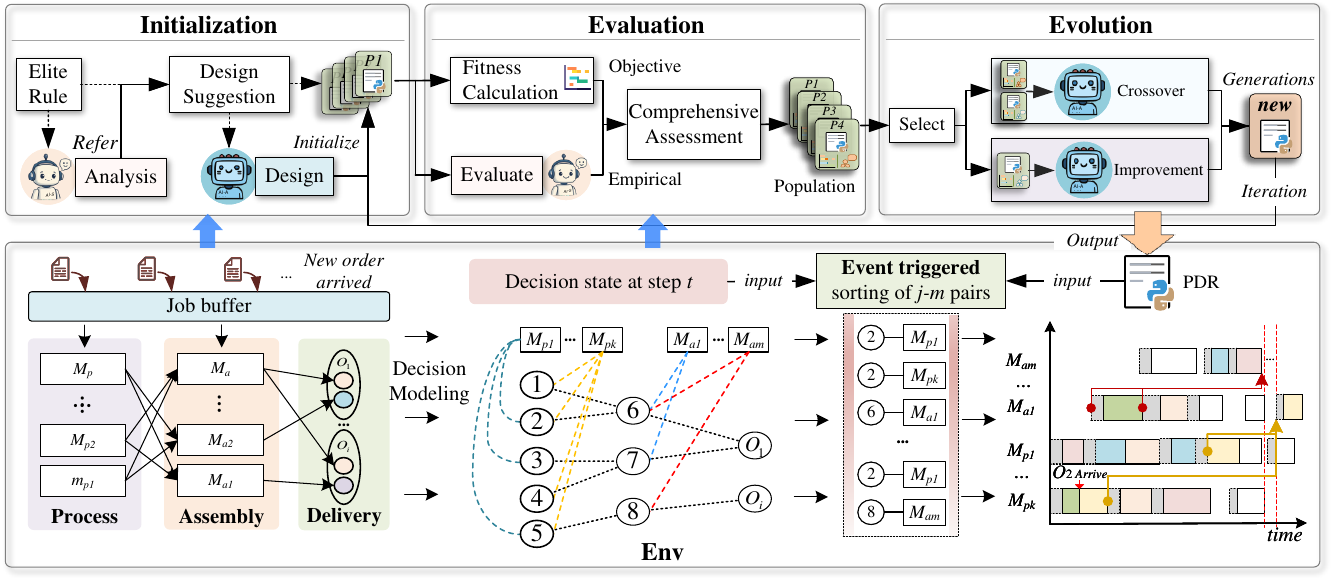}}
\caption{Automated dynamic PDR design based on EvoDR framework for FAFSP. By leveraging LLM to design a PDR that adapts to state changes, it achieves rapid sorting of $j$-$m$ pairs to perform job sequencing and machine allocation.}
\label{figure_framework}
\end{figure*}

The proposed EvoDR framework for dynamic FAFSP is illustrated in Fig.~\ref{figure_framework}. Upon order arrival, orders are decomposed into a hierarchy of processing and assembly jobs within the operation buffers. To satisfy kitting requirements, assembly is triggered only upon the completion of all prerequisite processing jobs, while final delivery is gated by the completion of the entire product set. This integrated coordination is modeled as directed edge selection on a heterogeneous graph. At each event-triggered decision step, the system evaluates feasible $(j,m)$ arcs, representing joint machine assignment and job sequencing, using the evolved PDR to rank candidates based on the real state.

\subsection{Evolution of PDRs}
\label{sec:evolution_rule}
\begin{figure*}[h]
\centering
\makebox[\textwidth][c]{\includegraphics[width=1\textwidth]{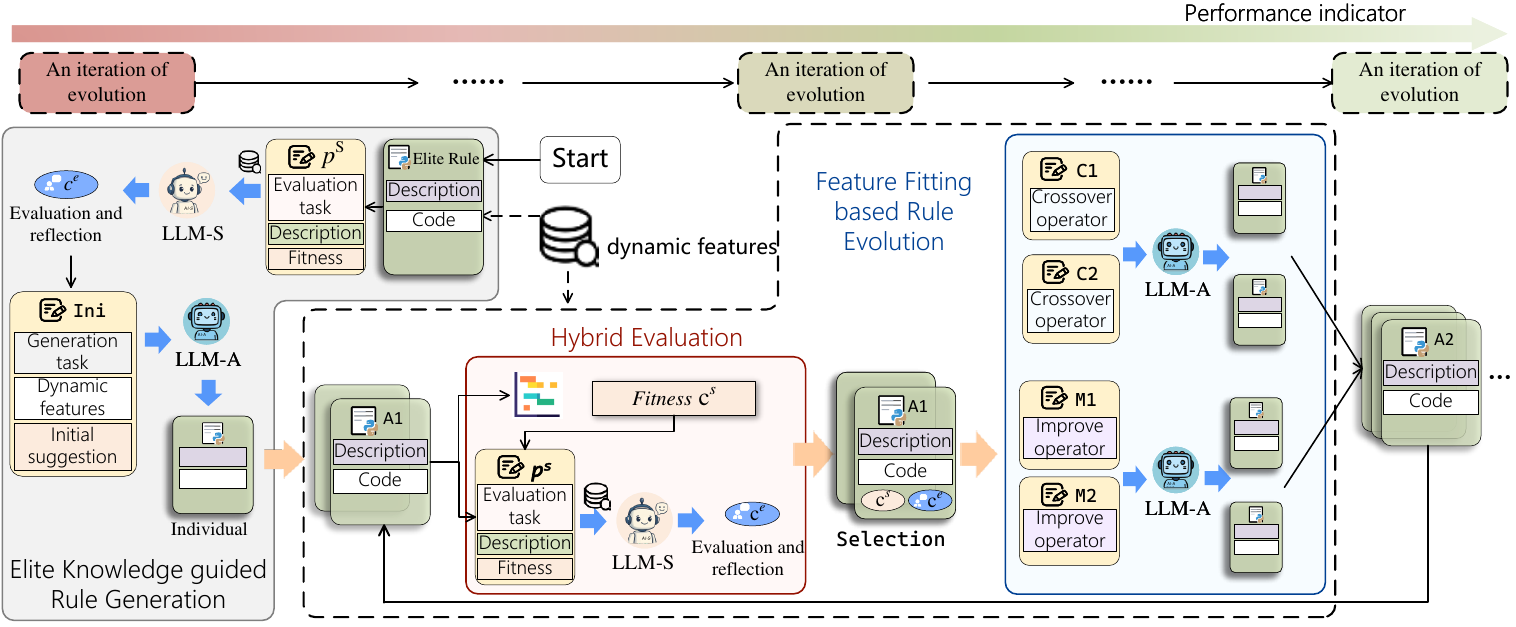}}
\caption{Automated design and evolution process for dynamic PDR. LLM-S introduces domain and elite knowledge to guide the initial rule initialization, and continuously guides the rule evolution through differentiated empirical evaluation of individuals.}
\label{figure_iteration_process}
\end{figure*}

The EvoDR framework is structured into four primary phases: encoding and prompt, initialization, evaluation, and evolution, shown in Fig~\ref{figure_iteration_process}. 
The evolution of PDRs is driven by the dual-expert mechanism. The algorithm expert (called LLM-A) $\mathbf{G}^{(A)}(\cdot)$ specializes in executable code generation of PDRs, while the scheduling expert (called LLM-S) $\mathbf{G}^{(S)}(\cdot)$ provides semantic analysis and reflective suggestions on heuristics. 
It instills high-level design knowledge into the population through an elite knowledge-guided initialization mechanism. 
During each iteration, each individual is assessed objectively by its fitness, while the semantic feedback from LLM-S guides the generation of subsequent rules.
This collaborative process facilitates the co-evolution of design knowledge and heuristic rules, exhibiting strong generalization across volatile production environments.

\begin{figure}[b]
\centering
\makebox[\columnwidth][c]{\includegraphics[width=1\columnwidth]{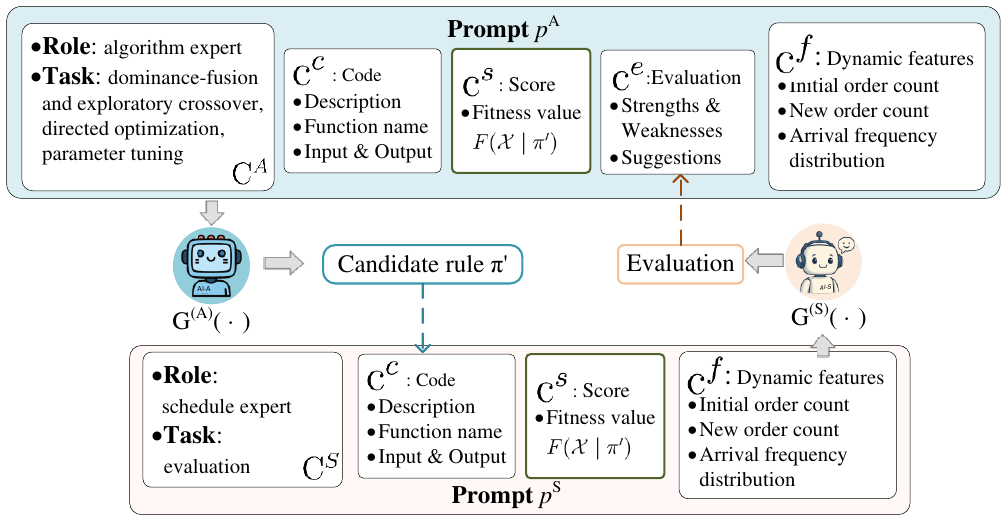}}
\caption{Design strategies and prompt engineering.}
\label{figure_prompt_struct}
\end{figure}

The PDR $\pi$ for solving this problem belongs to a specific algorithmic space $\mathbf{\Pi}$ generated by $\mathbf{G}^{(A)}(\cdot \mid p)$.
The LLM-A generates the candidate dispatching rules based on a given prompt $ p $, which includes disturbance features, problem descriptions, and existing rule design knowledge, etc.
The LLM-S provides feedback to supplement or adjust the $\pi'$ design suggestions, limitations, and other information to update the new prompt $ p'$ for optimizing the next iteration, as Eq.~\ref{eq_newprompt}:

\begin{equation}
\label{eq_newprompt}
p' = \mathbf{G}^{(S)}(\cdot \mid (\pi, F( \mathrm{x} \mid \pi))),\forall \mathrm{x} \in \mathcal{X}.
\end{equation}


\subsection{Prompt and Operator}
\label{sec:prompt}
The LLM-A and LLM-S undertake distinct analytical and design tasks, with structured prompt strategies effectively defining requirements. As illustrated in Fig.~\ref{figure_prompt_struct}, $C^{A}$ and $C^{S}$ denote the task descriptions embedded in the generation and evaluation prompts, respectively; $c^{f}$ the dynamic features of the scheduling state; $c^{c}$ the executable rule code; $c^{s}$ the objective fitness score; and $c^{e}$ the semantic evaluation returned by LLM-S. 
EvoDR employs two distinct prompts, $p^S$ and $p^A$, to drive the $\mathbf{G}^{(S)}$ and $\mathbf{G}^{(A)}$ respectively. 
$p^S$ integrates information such as the current individual's code $\mathrm{c}^c$, its objective fitness (the average performance across multiple instances, $F(\mathcal{X}|\pi)$), and the dynamic features of the scheduling state $\mathrm{c}^f$.
Furthermore, to generate insightful evaluation feedback $\mathrm{c}^e$, the prompt additionally provides the performance and descriptions of the “best” and “worst” rules discovered within the population. It instructs the LLM-S to \textit{``evaluate the following heuristic rules based on the above information''}, thereby forcing the model to identify specific strengths and weaknesses relative to the historical performance baselines rather than evaluating in isolation.

$p^A$ integrates the evaluation feedback $\mathrm{c}^e$ from $p^S$ with specific evolutionary operator instructions $\mathrm{C}^A$, driving the generation of new PDRs. By manipulating the different task descriptions in the prompt, we implement four evolutionary operators: \textbf{C1: Dominance-Fusion Crossover} establishes a context involving two parent PDRs and instructs the LLM: \textit{``Please design a new algorithm by combining the advantages of the two provided algorithms''}. This encourages the model to merge the structural strengths of parent PDRs into an excellent offspring. \textbf{C2: Exploratory Crossover} forces the LLM to design a novel PDR, preventing premature convergence and escaping local optima.
It provides parent code but explicitly states: \textit{``Please help me create a new algorithm that is completely different in form from the given one''}. This instruction prevents the LLM from simply mimicking the parent code, triggering the generation of new structures. \textbf{M1: Directed Optimization} aims to locally refine algorithms by focusing on logical refinement. It leverages the explicit suggestion field within feedback of empirical evaluation $c^e$. Prompts feed parent code and suggestions into the generator while specifying \textit{``Please improve the given algorithm based on the suggested enhancements''}. This translates natural language suggestions into code-level optimizations. \textbf{M2: Parameter Tuning} focuses on numerical refinement rather than structural change. The prompt instructs the LLM to \textit{``identify the main algorithm parameters, then create a new algorithm that has different parameter settings of the score function provided''}. It guides the LLM to perform fine-grained adjustments to weights and thresholds within an existing logic structure.



\subsection{Elite Knowledge Guided Initialization}
\label{sec:generation}
In traditional design methods such as GP, initial populations are often generated randomly to enhance diversity, but random terminal combinations are inefficient and difficult to extract semantic information. 
To overcome these limitations, this study proposes an elite knowledge guided rule generation mechanism, whose complete procedure is detailed in the Appendix. 
Leveraging the elite rule $\pi^0$, LLM-A and LLM-S collaborate to construct a structurally diverse and robust initial PDR population $P^0$, thereby establishing a high-quality starting point for the subsequent co-optimization process.

During the initialization phase, the $\mathbf{G}^{(A)}(\cdot)$, $\mathbf{G}^{(S)}(\cdot)$, and scheduling simulator $\mathbf{E}(\cdot)$ are initialized based on the instance set $\mathcal{X}$, and an elite rule $\pi^0$ is selected as the initial guiding knowledge.
The rule is executed in the simulator to obtain its performance $F(\mathcal{X}\mid \pi^0)$. 
The dynamic features $\mathrm{c}^f$, executable code $\mathrm{c}^c$, and performance score $\mathrm{c}^s$ are extracted to construct the evaluation prompt $p^S = (\mathrm{C}^S, \mathrm{c}^f, \mathrm{c}^c, \mathrm{c}^s)$.
Then, the $\mathbf{G}^{(S)}(\cdot)$ outputs an evaluation $\mathrm{c}^e$ of the elite rule based on $p^S$, summarizing the advantages and limitations of $\pi^0$ in the current scheduling environment and providing actionable improvement directions. The $\mathrm{c}^e$ is explicitly incorporated into the generation prompt for the algorithm generator, thereby injecting domain knowledge directly into the construction of new rule individuals.

The population size is $\mathcal{P}$, and the LLM sampling temperatures are set by sampling at equal intervals over a given range $(U_{low},U_{up})$, ensuring that early generations favor conservative structures while later generations enhance diversity. 
The generation prompt $p^A = (\mathrm{C}^A, \mathrm{c}^f, \mathrm{c}^c, \mathrm{c}^s, \mathrm{c}^e)$ is constructed and input into $\mathbf{G}^{(A)}(\cdot \mid (p^A, u))$ to generate a candidate rule $\pi'$. If $\pi'$ satisfies syntax, function interface, and scheduling semantics requirements, its performance $F(\mathcal{X}\mid\pi')$ is evaluated by the simulator, and the individual $(\pi', F(\mathcal{X}\mid\pi'))$ is added to the initial population $P^0$. 
Then, the differences between $\pi'$ and existing rules will be checked to avoid redundancy and enhance semantic diversity. 
When $F(\mathcal{X}\mid\pi')$ is better than the current best $F_{best}$, the corresponding temperature is recorded as $u_0$ for subsequent temperature control strategies. 

Through this elite knowledge guided mechanism, the initial population not only preserves the core scheduling logic of the elite rule $\pi^0$ but also incorporates the improvement directions revealed by the expert evaluation $\mathrm{c}^e$, resulting in a rule set that is both high-performing and structurally diverse. 

\subsection{Hybrid Evaluation}
\label{sec:hybrid_evaluation}
In scheduling rule evaluation, individuals are typically assessed only by the objective value from applying the rule to test instances. However, this single quantitative metric fails to characterize rule applicability in dynamic multi-stage environments or reveal latent strengths, such as superior performance under specific disturbances, and provides little actionable guidance for evolution. To address these limitations, a hybrid evaluation mechanism is proposed that comprises two independent channels. The objective channel computes each rule's fitness through the simulator, and this fitness is the sole basis for updating the best individual, selecting parents, and survival selection. The semantic channel, produced by LLM-S, provides structured feedback that is injected into the generation prompt to guide the generation of offspring rules, without affecting fitness computation or individual selection.


Each newly generated rule $\pi'$ is first evaluated objectively by the simulator $\mathbf{E}(\mathcal{X} \mid \pi')$ to calculate performance indicators. 
If the rule fails to execute, it is deemed invalid and removed; otherwise, it proceeds to the experiential evaluation stage. 
Then, LLM-S produces evaluation information $\mathrm{c}^e$, incorporating the rule’s textual description $\mathrm{C}^S$, dynamic features $\mathrm{c}^f$, code features $\mathrm{c}^c$, and initial scores $\mathrm{c}^s$. 
This is a comprehensive report, detailing the advantages, limitations, and specific improvement suggestions for the rule. This hybrid evaluation allows a holistic understanding of rule characteristics, providing clear guidance for subsequent rule generation and evolutionary refinement.


\subsection{Co-evolution of Design Knowledge and Rules}
\label{Rule_evolution}

In addition to the executable code, the design thoughts of PDR are also evolved. Furthermore, the competitiveness of the dynamic scheduling rules is reflected in their real-time response to the state. Therefore, a feature fitting rule evolution mechanism is proposed, using the improved suggestions from the hybrid evaluation to guide the evolution of the PDR population, as detailed in Algorithm~\ref{alg:rules_evolution}.

\begin{algorithm}[!t]
\caption{Co-evolution of design knowledge and rules}
\label{alg:rules_evolution}
\begin{algorithmic}[1]
    \STATE \textbf{Input:}
    Maximum number of iterations $K$, instance set $\mathcal{X}$, initial PDR population $P^0$, initial optimal PDR $\pi^*$, temperature $u_0$, dynamic scheduling simulator $\mathbf{E}(\cdot)$, neighborhood strategies $\mathcal{N}$, dynamic features $\mathrm{c}^f$, algorithm generator $\mathbf{G}^{(A)}(\cdot)$ and schedule evaluator $\mathbf{G}^{(S)}(\cdot)$.
    \STATE \textbf{Output:}
    The optimal PDR $\pi^*$ in $P$.
    \STATE Initialize the prompt elements $(\mathrm{C}^A/\mathrm{C}^S,\mathrm{c}^f,\mathrm{c}^c,\mathrm{c}^s,\mathrm{c}^e)$;
    \STATE $P,F_{best} \leftarrow P^0,V$;
    \FOR{$k=1,\ldots,K$}
    \STATE $OP_k \leftarrow \emptyset$; \# Initialize the offspring set
    \FOR{$n \in \mathcal{N}$}  
    \STATE $\mathrm{C}^{(A)} \leftarrow$ Get the task description of strategy $n$;
    \STATE $j \leftarrow$ The number of parents needed for strategy $n$;
    \STATE $PS \leftarrow$ Select $j$ rules in $P$ by TournamentSelection;
    \STATE $\mathrm{c}^c,\mathrm{c}^s \leftarrow$ Get the code and scores of parents in $PS$;
    \IF{whether evaluation is required in $n$}
    \STATE $p^S \leftarrow (\mathrm{C}^S,\mathrm{c}^f,\mathrm{c}^c,\mathrm{c}^s)$;
    \STATE $\mathrm{c}^e \leftarrow \mathbf{G}^{(S)}(\cdot \mid p^S)$;
    \STATE $p^A \leftarrow (\mathrm{C}^A,\mathrm{c}^f,\mathrm{c}^c,\mathrm{c}^s,\mathrm{c}^e)$;
    \ELSE
    \STATE $p^A \leftarrow (\mathrm{C}^A,\mathrm{c}^f,\mathrm{c}^c,\mathrm{c}^s)$;
    \ENDIF
    \STATE $\pi' \leftarrow \mathbf{G}^{(A)}(\cdot \mid (p^A,u_0))$;
    \IF{$\pi'$ is valid}
    \STATE $F(\mathcal{X} \mid \pi') \leftarrow \mathbf{E}(\mathcal{X} \mid \pi')$;
    \STATE $OP_{k} = OP_{k} \cup(\pi',F(\mathcal{X} \mid \pi'))$;
    \IF{$F(\mathcal{X} \mid \pi') \leq F_{best}$}
    \STATE $\pi^*, F_{best} \leftarrow \pi',F(\mathcal{X} \mid \pi')$; 
    \ENDIF
    \ENDIF
    \ENDFOR
    \STATE $P \leftarrow $ EliteSelection($P \cup OP_{k}$); \# Survival selection
    \ENDFOR
    
\end{algorithmic}
\end{algorithm}

During co-evolution, different evolutionary operators (C1, C2, M1 and M2 in Fig.~\ref{figure_iteration_process} ) will be used one by one as neighborhood strategies $n \in \mathcal{N}$ for the PDR population. Parent individuals are first selected via Tournament Selection strategy to form the set $PS$. Their code representations and scheduling performance are extracted to form $(\mathrm{c}^c, \mathrm{c}^s)$. Depending on whether semantic evaluation is required, a prompt $p^S$ is constructed as $(\mathrm{C}^S,\mathrm{c}^f,\mathrm{c}^c,\mathrm{c}^s)$ and the evaluator $\mathbf{G}^{(S)}$ produces the semantic assessment result $c^e$. 
Specifically, during the evolution process, $c^e$ records the structured design knowledge of different PDRs, including: 1) the rule’s advantageous features in scheduling; 2) the summary of its potential limitations; and 3) improvement suggestions. 
This semantic information, along with numerical performance, is input to guide the direction of subsequent rule generation.
The generation prompt $p^A$ is then constructed as $(\mathrm{C}^A,\mathrm{c}^f,\mathrm{c}^c,\mathrm{c}^s,\mathrm{c}^e)$ and the $\mathbf{G}^{(A)}$ produces the offspring rule under temperature $u_0$ as Eq.~\ref{eq2_1}.

\begin{equation}
\label{eq2_1}
\pi' \sim \mathbf{G}^{(A)}(\cdot \mid (p^A, u_0))
\end{equation}

The crossover operators vary in parental usage of $\mathbf{G}^{(S)}$: C1 leverages both parents' traits for deep integration, while C2 relaxes constraints to reduce dependency. Improvement operators similarly differ, with M1 using external guidance and M2 focusing on internal structures, enabling a balance between intensive exploitation and novelty exploration.
The generated offspring rule $\pi'$ undergoes feasibility checking and scheduling performance evaluation as $F(\mathcal{X} \mid \pi')$. 
The evaluation results are used to update the best rule $\pi^*$ and the next-generation population $OP_k$. Parent selection ensures that high-quality rules influence subsequent generations, while temperature $u_0$ controls exploration to maintain a balance between performance and diversity.

Through this mechanism, crossover and improvement operations are no longer governed by static operators but are dynamically driven by the interaction among prompts, semantic evaluation, and the generator. This enables continuous interplay and mutual reinforcement between domain knowledge and rule structures throughout the iterative cycles. Consequently, the rule population evolves under the joint influence of numerical fitness and semantic knowledge, yielding scheduling rules with strong and stable performance across the tested scenarios.

\section{Experimental Results and Analysis}
\label{Exp:experience}
\subsection{Dataset and Implementation}
\label{Exp:implementation}
To the best of our knowledge, no publicly available benchmark exists for the dynamic multi-product delivery FAFSP with dual kitting constraints. 
Consequently, the experimental evaluation in this study is conducted on a collection of test datasets that integrate real-world industrial instances and synthetically generated instances. 
The real-world instances originate from the order and resource configuration data of a Chinese home-appliance manufacturer within a specific period, including information on order structure, machine resources, and order arrival distribution. The generated instances follow the settings commonly adopted in existing dynamic scheduling studies~\cite{basir2018bi,lei2023large,qiu2024multi} for configuring order arrival processes.
Each operating condition is specified by the machine configuration $m_1$ (processing) and $m_2$ (assembly), the flexibility factor $\varphi$, the number of dynamically arriving orders $\alpha$, and the load factor $\mu$.

Experimental settings are detailed in the corresponding subsections. Notably, all classical and LLM-generated PDRs exhibit millisecond-level decision times, fully satisfying the real-time requirements of dynamic FAFSP. EvoDR and comparative algorithms are implemented in Python\footnote{https://github.com/jhqiu1/EvoDR} and executed on an Intel Core i5-12600KF processor.

\subsection{Parameters Setting and Analysis}
\label{Exp:Hyperparametric}
The EvoDR's parameters include the number of iterations, population size, LLMs, and corresponding temperature settings. 
Following existing studies~\cite{EOH}, the number of iterations is set as 20 in all experiments, and the population size is set to 4-6. 
Specifically, in the comparison of AHD methods in Section~\ref{Exp:Comparison_AHDs}, the population size is extended to 6 in order to compare the performance of different AHD methods in detail and to limit the maximum total sampling budget. 

\textbf{Different LLMs:}
Well-known LLMs were selected to evaluate the differences in the quality and cost of generating rules.
Therefore, experimental comparisons are conducted across 20 instances under the operating condition (3-6-0.5-20-1), covering all instances $T_{all}$, randomly divided training sets $T_{train}$, and test sets $T_{test}$. The Ratio is calculated as the improvement of the best designed PDR over the earliest due date (EDD) divided by the corresponding cost.
All results are presented in Table~\ref{tab: LLMs}.

\begin{table}[h]
\caption{Comparison of different LLMs.}\label{tab: LLMs}
\centering
\resizebox{1\columnwidth}{!}{
\begin{threeparttable} 
\begin{tabular}{lllllcll}
\toprule
LLM     & \textbf{$T_{train}$} & \textbf{$T_{test}$} & \textbf{$T_{all}$}  & Token & Acc & Cost & Ratio    \\
\midrule
QW8\dag           & 1244.02          & 1052.85         & 1148.44         & 0.37M          & 80.5\%              & 0.40          & 0.000647          \\
QW235\dag         & \textbf{1015.66} & 951.93          & 983.79          & 0.38M          & 94.5\%              & 1.67          & 0.002133          \\
deepseek chat & 1090.51          & 1074.69         & 1082.61         & \textbf{0.28M} & 91.5\%              & 0.65          & 0.000950          \\
claude        & 1021.45          & \textbf{896.38} & \textbf{958.91} & 0.53M          & 75.0\%              & 5.77          & 0.007144          \\
gpt-4o-mini   & 1082.17          & 974.07          & 1028.12         & 0.33M          & \textbf{96.0\%}     & \textbf{0.09} & \textbf{0.000122}\\
\bottomrule
\end{tabular}
\begin{tablenotes}    
    \footnotesize            
   \item  $\dag$: the Open source model. Token: the total consumed tokens. Acc: the accuracy of the generated code. Cost: the total expense consumed using the commercial interface. Ratio: the cost performance ratio. Bold numbers indicate the best performance.
\end{tablenotes}      
\end{threeparttable}}
\end{table}

The results indicate that changing the underlying model significantly affects both code accuracy and cost. Although QW235 and Claude achieve competitive performance on different sets of instances, both show disadvantages in accuracy and total expenditure. 
Considering the overall cost and accuracy, GPT-4o-mini is selected for all subsequent experiments.

\textbf{Temperature setting:} the default range is $(U_{low},U_{up})=(0.3,1.5)$; a sensitivity analysis over five representative ranges and the corresponding results are provided in the Appendix.

\textbf{Cost analysis:} a full EvoDR design run costs about \$0.189 (\$0.123 for LLM-A and \$0.066 for LLM-S); the offline design cost is one-time, while online dispatching is sub-millisecond and independent of LLM calls; detailed cost comparisons are given in the Appendix.

\subsection{Ablation Experiment}
\label{Exp:Ablation}
The effectiveness of each improvement component is discussed in this section. 
The ablated baseline of EvoDR is denoted as EDR, and each proposed module is removed independently. 
To assess the influence of dynamic disturbance information, all dynamic features are excluded from the prompts while the remaining procedure is kept unchanged, which is referred to as EDR-D. 
Removing the elite knowledge guided rule generator from the initialization stage results in EDR-E. 
As crossover and mutation prompts act as essential evolutionary operators, their removal produces EDR-C and EDR-M, respectively.
Furthermore, we remove the semantic evaluation component introduced by LLM-S from the hybrid evaluation, and compare it with methods using fixed improvement suggestions and random selection from the suggestion set.
Additionally, the feature designs for orders, machines, and arcs are extracted separately to investigate the sensitivity of EvoDR to these features.
The experimental settings follow those presented in Sections~\ref{Exp:implementation} and~\ref{Exp:Hyperparametric}, and testing is conducted on 20 instances generated based on real-world industrial features. 
All methods were executed independently three times, and the number of iterations in all LLM-based methods is 20 and the population size is 6. The results are summarized in Table~\ref{tab: ablation_experiment}.

\begin{table}[h]
\caption{Results of ablation experiment.}
\label{tab: ablation_experiment}
\centering
\begin{adjustbox}{center}
\resizebox{1\linewidth}{!}{
\begin{tabular}{lcccc}
\toprule
\textbf{Method} & \textbf{Max} & \textbf{Min} & \textbf{Avg.}  & \textbf{ARI$\uparrow$} \\ \midrule
EDR-D           & 1242.86      & 1134.08      & 1195.37±55.69  & 6.08\%       \\
EDR-E           & 1681.21      & 1112.70      & 1434.77±291.70 & -12.73\%     \\
EDR-C           & 1183.04      & 1117.04      & 1151.20±33.06  & 9.55\%       \\
EDR-M           & 1257.91      & 1087.66      & 1185.85±88.08  & 6.83\%       \\\midrule
EDR-F(order)    & 1710.45      & 1316.90      & 1579.27±227.22 & -24.08\%     \\
EDR-F(machine)  & 1215.98      & 1101.01      & 1140.60±65.30  & 10.39\%      \\
EDR-F(arc)      & 1663.30      & 1566.91      & 1630.28±54.90  & -28.09\%     \\ \midrule
EDR-S           & 1333.00      & 1234.78      & 1272.80±52.73  & 0.00\%       \\
EDR-S(fix)      & 1333.40      & 1156.19      & 1215.87±101.79 & 4.47\%       \\
EDR-S(random)   & 1116.78      & 1101.04      & 1107.07±8.49   & 13.02\%      \\ \midrule
EvoDR           & \textbf{1097.23}      & \textbf{1072.17}      & \textbf{1084.17±12.56}  & \textbf{14.82\%}     \\
\bottomrule
\end{tabular}}
\end{adjustbox}
\end{table}

EvoDR achieves the best performance with the lowest average tardiness (1084.17) and highest ARI (14.82\%),  validating that all proposed modules collectively improve its optimization efficiency. Component ablation further reveals distinct roles: removing the elite module (EDR-E) raises the average tardiness to 1434.77, confirming the value of high-quality initialization; removing the crossover (C1, C2) or improvement (M1, M2) operators degrades performance, indicating their complementary roles in exploration and exploitation; and removing dynamic information (EDR-D) increases tardiness to 1195.37, underscoring the value of global environment information.

Interestingly, evaluating the hybrid evaluation mechanism yields unexpected insights. Removing the LLM-S semantic evaluation entirely (EDR-S) degrades tardiness to 1272.80. However, substituting it with fixed (1215.87) or random (1107.07) suggestions performs better. This suggests that providing clear feedback can effectively broaden the search scope of LLMs and prevent them from converging prematurely, even without semantic optimization. 

\subsection{Comparative Experiment} 
\label{Exp:Comparison_AHDs}

\begin{table*}[t]
\caption{Comparison of different AHD methods.}
\label{tab: compara_AHD_experiment}
\centering
\begin{adjustbox}{center}
\resizebox{1\textwidth}{!}{
\begin{threeparttable}
\begin{tabular}{cccccccccc}
\toprule
\textbf{Inst.} & \textbf{EDD} & \textbf{GP}      & \textbf{GEP}      & \textbf{RandSample} & \textbf{FunSearch} & \textbf{EoH}     & \textbf{MCTS-AHD} & \textbf{ReEvo}   & \textbf{EvoDR}  \\
\midrule
1                                     & 18620.87     & 13779.29$\pm$1583.03        & \textbf{12888.94$\pm$676.57} & 16125.55$\pm$1072.87        & 15501.02$\pm$1276.06        & 16606.39$\pm$188.84          & 16025.71$\pm$2511.83        & 16269.56$\pm$772.33         & 14918.53$\pm$738.55           \\
2                                     & 6209.37      & 5206.79$\pm$171.79          & 5439.67$\pm$118.26           & 5074.58$\pm$178.02          & 5180.63$\pm$194.74          & 4997.67$\pm$247.11           & 5128.14$\pm$323.62          & 5025.17$\pm$82.26           & \textbf{4983.60$\pm$123.33}   \\
3                                     & 14040.38     & 15286.23$\pm$620.67         & 15889.09$\pm$751.30          & 14666.24$\pm$518.11         & 14920.71$\pm$500.68         & 15506.72$\pm$1690.63         & 15915.30$\pm$1628.90        & 15948.22$\pm$2414.32        & \textbf{13718.30$\pm$1322.11} \\
4                                     & 5615.83      & 4508.49$\pm$172.44          & 4494.34$\pm$82.72            & 4591.96$\pm$274.80          & 4677.21$\pm$167.20          & \textbf{4397.28$\pm$140.58}  & 4456.35$\pm$111.37          & 4625.54$\pm$198.97          & 4427.48$\pm$183.37            \\
5                                     & 8564.67      & 7129.13$\pm$844.62          & 7914.79$\pm$572.44           & 6498.42$\pm$851.99          & 7259.66$\pm$581.37          & 7005.32$\pm$512.93           & 7745.51$\pm$1564.11         & 7586.26$\pm$1606.54         & \textbf{6176.54$\pm$270.17}   \\
6                                     & 5401.75      & 3582.52$\pm$200.21          & 3771.46$\pm$190.15           & 3692.99$\pm$199.03          & 3856.92$\pm$213.78          & 3800.25$\pm$244.37           & 4026.00$\pm$370.74          & 3890.38$\pm$332.11          & \textbf{3471.25$\pm$156.75}   \\
7                                     & 5306.45      & \textbf{4886.91$\pm$487.79} & 5701.90$\pm$551.75           & 5212.66$\pm$299.20          & 5246.81$\pm$434.28          & 5077.85$\pm$368.14           & 5634.60$\pm$249.19          & 5355.93$\pm$359.83          & 4958.29$\pm$388.79            \\
8                                     & 3113.85      & 2619.01$\pm$196.23          & 2716.35$\pm$99.73            & 2444.02$\pm$238.98          & \textbf{2413.26$\pm$130.94} & 2426.47$\pm$101.31           & 2434.68$\pm$204.44          & 2584.65$\pm$317.76          & 2421.06$\pm$49.40             \\
9                                     & 15820.28     & 16733.99$\pm$1786.72        & 17773.62$\pm$1165.02         & 15335.59$\pm$371.19         & 15147.89$\pm$759.68         & \textbf{14181.08$\pm$777.92} & 14345.73$\pm$710.97         & 14586.37$\pm$446.35         & 15066.96$\pm$712.06           \\
10                                    & 6398.62      & 5645.15$\pm$715.20          & 6595.46$\pm$361.02           & 5042.04$\pm$244.70          & 5178.99$\pm$202.42          & 6073.79$\pm$594.05           & 5576.02$\pm$856.01          & 5624.99$\pm$715.46          & \textbf{4889.82$\pm$290.04}   \\
\midrule
11                                    & 3150.55      & 3482.09$\pm$705.00          & 4013.67$\pm$387.63           & 2937.59$\pm$160.20          & 2866.28$\pm$106.43          & 2877.80$\pm$97.69            & 3049.71$\pm$245.76          & 2978.32$\pm$106.57          & \textbf{2852.79$\pm$78.30}    \\
12                                    & 4689.78      & 4144.37$\pm$639.18          & 4646.20$\pm$306.09           & 3770.40$\pm$86.07           & 3778.30$\pm$194.73          & \textbf{3761.24$\pm$102.24}  & 3879.92$\pm$116.97          & 4008.66$\pm$241.41          & 3873.35$\pm$167.08            \\
13                                    & 4028.45      & 5614.78$\pm$2235.00         & 6623.13$\pm$1159.47          & 3728.43$\pm$206.27          & 3596.64$\pm$324.88          & \textbf{3400.84$\pm$206.78}  & 4440.97$\pm$1464.87         & 4012.66$\pm$495.46          & 3659.26$\pm$397.92            \\
14                                    & 6514.52      & 6787.40$\pm$1972.00         & 8648.84$\pm$926.04           & \textbf{5572.10$\pm$627.87} & 5920.40$\pm$682.68          & 6606.24$\pm$258.01           & 6361.14$\pm$746.49          & 6560.46$\pm$1341.46         & 5580.81$\pm$125.47            \\
15                                    & 4713.58      & 4384.48$\pm$546.86          & 4413.90$\pm$499.63           & 4331.87$\pm$633.85          & 4089.39$\pm$610.40          & 4267.29$\pm$723.96           & \textbf{4007.20$\pm$520.46} & 4126.82$\pm$529.14          & 4036.83$\pm$469.30            \\
16                                    & 6737.87      & 4182.07$\pm$393.73          & \textbf{4096.37$\pm$418.56}  & 4367.37$\pm$449.91          & 4122.71$\pm$348.62          & 4645.99$\pm$397.15           & 4613.52$\pm$739.17          & 4708.95$\pm$485.15          & 4647.74$\pm$496.72            \\
17                                    & 3703.18      & 3580.72$\pm$785.35          & 3972.45$\pm$576.08           & 3044.10$\pm$65.45           & 3055.83$\pm$283.83          & \textbf{2903.73$\pm$350.81}  & 3352.39$\pm$514.33          & 3297.65$\pm$374.09          & 2917.49$\pm$175.23            \\
18                                    & 7465.48      & 5719.70$\pm$561.44          & 5780.75$\pm$886.59           & 6147.06$\pm$388.24          & 6054.04$\pm$657.44          & 5706.07$\pm$296.73           & \textbf{5467.42$\pm$465.49} & 5954.54$\pm$919.65          & 5812.73$\pm$361.15            \\
19                                    & 4701.47      & \textbf{3161.73$\pm$440.12} & 3559.27$\pm$168.68           & 3312.39$\pm$248.78          & 3259.23$\pm$209.67          & 3216.57$\pm$286.64           & 3395.24$\pm$266.54          & 3617.67$\pm$495.06          & 3269.54$\pm$279.56            \\
20                                    & 4349.80      & 4153.66$\pm$324.39          & 4050.97$\pm$113.20           & 4302.82$\pm$235.88          & 4043.24$\pm$128.26          & 4030.14$\pm$228.55           & 4275.04$\pm$572.39          & \textbf{4007.00$\pm$207.92} & 4079.78$\pm$179.66            \\
\midrule
All(Avg.)                                  & 6957.34      & 6229.43                     & 6649.56                      & 6009.91                     & 6008.46                     & 6074.44                      & 6206.53                     & 6238.49                     & \textbf{5788.11}              \\
ARI$\uparrow$                                 & 0.00\%       & 10.46\%                     & 4.42\%                       & 13.62\%                     & 13.64\%                     & 12.69\%                      & 10.79\%                     & 10.33\%                     & \textbf{16.81\%}              \\
\midrule
Train(Avg.)                                 & 8909.21      & 7937.75                     & 8318.56                      & 7868.41                     & 7938.31                     & 8007.28                      & 8128.80                     & 8149.71                     & \textbf{7503.18}              \\
ARI$\uparrow$                                 & 0.00\%       & 10.90\%                     & 6.63\%                       & 11.68\%                     & 10.90\%                     & 10.12\%                      & 8.76\%                      & 8.52\%                      & \textbf{15.78\%}              \\
\midrule
Test(Avg.)                                  & 5005.47      & 4521.10                     & 4980.56                      & 4151.41                     & 4078.61                     & 4141.59                      & 4284.26                     & 4327.27                     & \textbf{4073.03}              \\
ARI$\uparrow$                                   & 0.00\%       & 9.68\%                      & 0.50\%                       & 17.06\%                     & 18.52\%                     & 17.26\%                      & 14.41\%                     & 13.55\%                     & \textbf{18.63\%}              \\
\midrule
Avg. Rank   & 7.85     & 5.00     & 6.35     & 4.25     & 3.90     & 3.35     & 4.95     & 5.55     & \textbf{2.65} \\
W / L       & 5 / 95     & 37 / 63    & 18 / 82    & 38 / 62    & 38 / 62    & 36 / 64    & 34 / 66    & 29 / 71    & —    \\
Wilcoxon $p$ & 1.35e-16 & 1.37e-04 & 1.11e-09 & 1.37e-03 & 2.51e-03 & 2.45e-03 & 6.65e-05 & 1.24e-05 & —   
\\
                 \bottomrule
\end{tabular}

\begin{tablenotes}    
    \footnotesize            
   \item  The design task is repeated five times for all AHD methods, and the mean performance and variance of the best PDR are recorded. Bold numbers indicate the best performance, and ARI is average relative improvement, computed as $(T_{max} - T_{current})/T_{max}$, where $T_{max}$ is the average tardiness of the worst-performing baseline and $T_{current}$ that of the current method. W / L refers to the number of winning and losing instances compared to the proposed method.
\end{tablenotes}      
\end{threeparttable}}
\end{adjustbox}
\end{table*}

This section compares EvoDR with traditional and advanced AHD methods, including traditional methods such as GP~\cite{zhang2020evolving} and GEP, and advanced LLM-based AHD methods such as FunSearch based on island-model exploration~\cite{funsearch}, EoH based on evolutionary search~\cite{EOH, meoh}, MCTS-AHD based on Monte Carlo tree search\cite{mcts_ahd}, ReEvo incorporating reflective reasoning~\cite{reevo}, and RandSample (random repeated prompting) to generate heuristics without an iterative search framework~\cite{Random}. 
To ensure fairness, the dynamic features used in EvoDR for the dynamic FAFSP are added to the terminal sets of GP and GEP. 
All LLM-based approaches adopt the same initial PDR setting and are permitted to access the same dynamic features. 
Each method is evaluated based on its convergence efficiency and the performance of the rules generated on the training set. Its generalization is then assessed on the test instance set.
In particular, 20 instances that correspond to the operating condition (3-6-0.5-20-2) with real industrial characteristics are selected, and 10 of them are randomly assigned as the training set. 
Each method is independently conducted for design tasks, and the maximum number of samples is set to 200 and the population size is 6 for the different methods.
The tardiness performance of all methods over all instances, the training instances, and the test instances are reported in Table~\ref{tab: compara_AHD_experiment}.

\begin{figure}[h]
\centering
\makebox[\columnwidth][c]{\includegraphics[width=1\columnwidth]{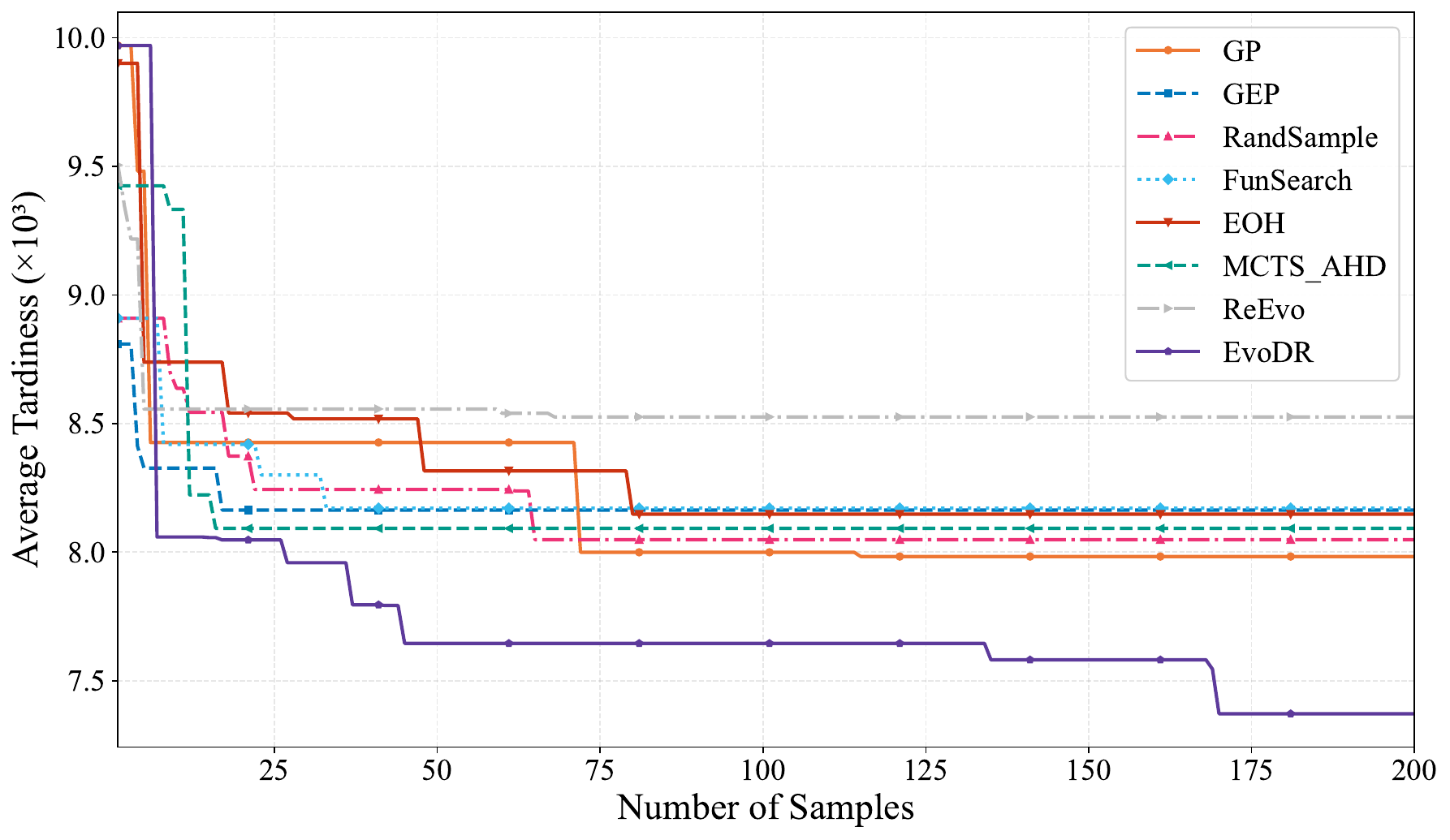}}
\caption{Convergence of different AHD methods.}
\label{figure_convergence_AHDs}
\end{figure}

The training results show that all AHD methods outperform the EDD baseline with an average tardiness of 8909.21. 
When the dynamic-feature terminal set is incorporated, GP achieves better performance than GEP, improving tardiness by approximately 10.90\% and reaching 7937.75. 
Among the LLM-based approaches, the RandSample attains the best performance, ranking second only to EvoDR, and EoH achieves an improvement of 10.12\%. 
In contrast, ReEvo, which integrates reflective reasoning, exhibits degraded performance with an average tardiness of 8149.71. FunSearch and MCTS-AHD, utilising island-model exploration and Monte Carlo tree search respectively, yield improvements of 10.90\% and 8.76\%.

\begin{figure*}[t]
\centering
\makebox[\textwidth][c]{\includegraphics[width=0.9\textwidth]{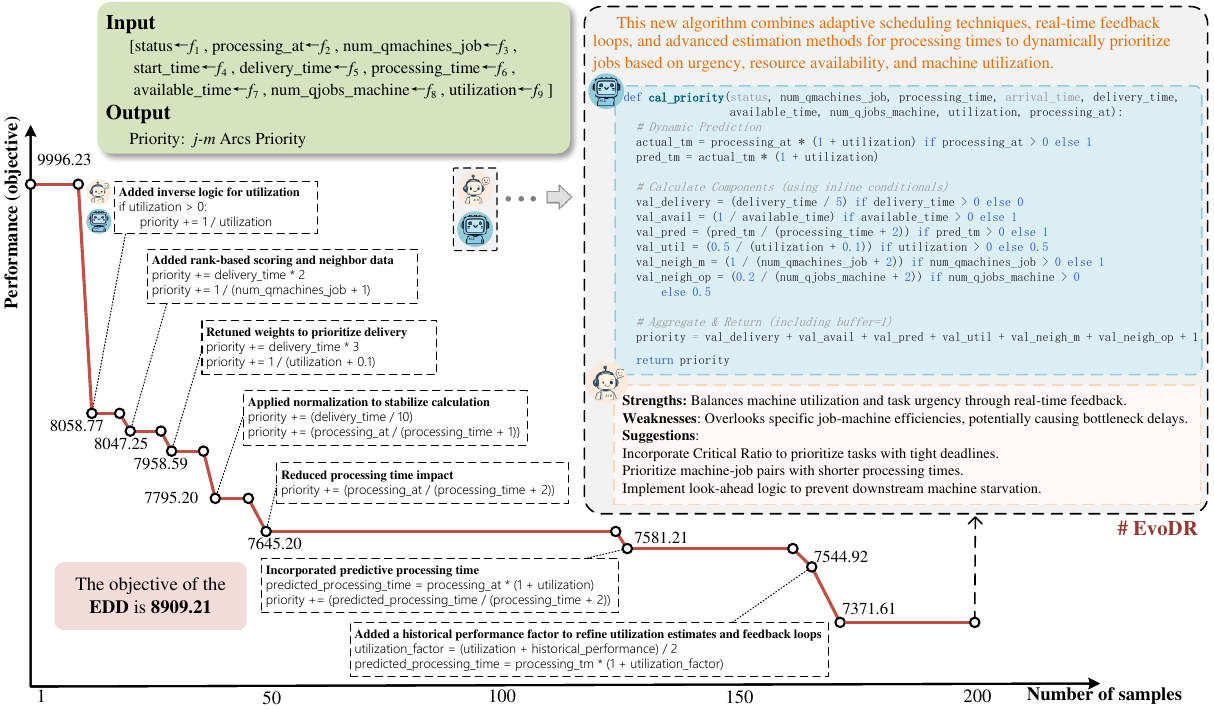}}
\caption{Convergence analysis for dynamic PDR evolution. }
\label{figure_convergence_analys}
\end{figure*}
On the test set, FunSearch exhibits slightly lower performance with an average tardiness of 4078.61.
Notably, GP fails to maintain its favourable training performance and shows substantial degradation on the test set, reaching 4521.10. 
Unlike methods whose performance appears to reflect overfitting to the training set, 
EvoDR benefits from a better understanding and utilization of problem-structure information, thereby generating rules with stronger generalisation ability.
Overall, EvoDR achieves the lowest average tardiness across the twenty instances, resulting in an average improvement of 16.81\% and improving upon EDD by 18.63\% on the test set.

Then, we conduct a Friedman test (significance level 0.05, 20 instances as blocking factors, 8 methods ranked per instance by mean tardiness) and two-tailed Wilcoxon signed-rank tests (100 paired observations per comparison, EvoDR as baseline, Bonferroni-corrected threshold $0.05/7 \approx  0.0071$). The Friedman test confirms significant differences among methods ($p < 0.0001$). EvoDR ranks first (Avg.Rank = 2.65), and Wilcoxon tests show it significantly outperforms all competitors (all $p \le 0.0026$, below the Bonferroni threshold).

The convergence curves of the generated PDRs in a single training design further illustrate the relative advantages of the AHD method, as shown in Fig.~\ref{figure_convergence_AHDs}.
With 200 samples, GP identifies a superior PDR within approximately 70 samples and achieves the best early-stage performance. 
GEP, RandSample, and FunSearch also produce strong initial individuals, but similar to other approaches, they converge prematurely around an average tardiness of $8.15 \times 10^{3}$, with limited improvement thereafter. 
In contrast, EvoDR, despite starting from weaker initial individuals, surpasses the current best performance on eight occasions within two hundred iterations and continues to improve after one hundred samples. 
These results demonstrate that EvoDR is able to incorporate experience and structural insights from the dynamic dual kitting FAFSP, enabling the continuous generation of high-quality PDRs.

\subsection{Iterative Process Analysis}
\label{Exp:Iterative_analysis}
This section provides an interpretability analysis of EvoDR's evolution, and the convergence process is illustrated in Fig.~\ref{figure_convergence_analys}. 
It can be observed that LLM-S plays a pivotal role in ensuring continuous improvement, functioning not merely as a selector but as a "semantic navigator" that guides the evolutionary search.
By analyzing the structural logic and performance feedback of candidate rules, LLM-S generates actionable textual suggestions, such as identifying weight imbalances or logical deficiencies to guide the algorithm generator LLM-A in performing optimization. This mechanism effectively bridges the gap between quantitative objective metrics and qualitative code structures, enabling the optimization process to resolve complex logical conflicts that numerical fitness alone cannot address.

In the first 50 samplings, the curve exhibits a rapid descent driven by structural exploration, establishing a strong baseline. Notably, even after the 150th sampling, a phase where traditional heuristics typically plateau (shown in Fig.~\ref{figure_convergence_AHDs}), the curve demonstrates a significant secondary convergence, dropping from an objective value of 7581.21 to 7371.61. 
This sustained late-stage improvement highlights the critical effectiveness of the Hybrid Evaluation module. By combining quantitative fitness metrics with qualitative expert analysis, the module identifies subtle semantic constraints, such as “\textit{Added a historical performance factor to refine utilization estimates and feedback loops}”, which cannot be revealed by numerical indicators alone. This deep semantic feedback provides the necessary guidance for the directed optimization operator to perform precise feature fitting, successfully injecting advanced dynamic features like predictive processing times to ensure consistent performance across conditions and break through local optima.

\subsection{Robustness Analysis} %
\label{Exp:Robustness}

\begin{table*}[t]
\caption{Results of robust experiment.}
\label{tab: robust_experiment}
\centering
\begin{adjustbox}{center}
\resizebox{1\textwidth}{!}{
\begin{tabular}{ccccccccccc}
\toprule
$\mathrm{m}$-$\varphi$-$\alpha$-$\mu$     & \textbf{EDD} & \textbf{FIFO+SPT} & \textbf{FIFO+EET} & \textbf{MOPNR+SPT} & \textbf{MOPNR+EET} & \textbf{LWKR+SPT} & \textbf{LWKR+EET} & \textbf{MWKR+SPT} & \textbf{MWKR+EET} & \textbf{EvoDR}    \\ 
\midrule
3-6-0.5-20-1                      & 1710.45      & 1729.20           & 2131.53           & 2029.53            & 2447.68            & 2219.19           & 2675.79           & 1847.57           & 2091.60           & \textbf{1117.80}  \\
3-6-0.5-20-2                      & 6957.34      & 7008.93           & 7312.52           & 7719.59            & 8055.02            & 7662.46           & 8114.47           & 7876.31           & 8205.37           & \textbf{6265.49}  \\
3-6-0.5-20-4                      & 13472.86     & 13668.39          & 13689.63          & 16854.99           & 16863.26           & 14953.49          & 15198.38          & 15232.15          & 15222.27          & \textbf{11079.05} \\
3-6-0.5-50-1                      & 4080.61      & 2678.10           & 3586.94           & 2913.22            & 3772.69            & 2889.80           & 3745.43           & 2731.95           & 3602.20           & \textbf{2693.39}  \\
3-6-0.5-50-2                      & 11796.76     & 10636.39          & 11618.96          & 11701.54           & 12690.06           & 11388.47          & 12222.28          & 10911.00          & 12084.94          & \textbf{8995.10}  \\
3-6-0.5-50-4                      & 30322.88     & 30122.07          & 30205.86          & 37851.62           & 38635.80           & 35150.07          & 34952.64          & 37470.82          & 37470.92          & \textbf{25278.60} \\
3-6-0.7-20-1                      & 1709.87      & 1701.79           & 2008.32           & 1909.75            & 2279.13            & 1844.23           & 2083.45           & 1697.93           & 1893.94           & \textbf{1035.62}  \\
3-6-0.7-20-2                      & 5500.23      & 5009.23           & 5307.16           & 6333.49            & 6497.30            & 6023.81           & 6309.04           & 5685.14           & 5808.45           & \textbf{4166.53}  \\
3-6-0.7-20-4                      & 13581.56     & 13344.90          & 13422.08          & 16056.94           & 16065.12           & 13881.40          & 14115.48          & 14180.19          & 14264.88          & \textbf{11308.20} \\
3-6-0.7-50-1                      & 3092.07      & \textbf{1698.89}  & 2602.44           & 1704.36            & 2612.33            & 1721.88           & 2583.25           & 1827.41           & 2701.33           & 1813.76           \\
3-6-0.7-50-2                      & 9761.05      & 7782.37           & 9005.45           & 7965.17            & 9164.06            & 7856.55           & 9023.75           & 7708.64           & 8996.94           & \textbf{7566.31}  \\
3-6-0.7-50-4                      & 31499.53     & 32061.02          & 32088.02          & 37289.87           & 37480.73           & 34693.13          & 34045.23          & 36257.71          & 35879.24          & \textbf{22108.98} \\
5-12-0.5-20-1                     & 942.39       & 269.27            & 866.83            & 269.05             & 865.22             & 303.33            & 884.05            & \textbf{230.79}   & 799.88            & 238.68            \\
5-12-0.5-20-2                     & 2980.40      & 1849.67           & 2807.27           & 1920.71            & 2836.39            & 2039.29           & 3055.92           & 1738.17           & 2690.81           & \textbf{1670.05}  \\
5-12-0.5-20-4                     & 2766.35      & 2664.84           & 2966.33           & 2770.40            & 3233.97            & 2893.50           & 3270.72           & 2489.23           & 3000.27           & \textbf{1989.17}  \\
5-12-0.5-50-1                     & 2353.79      & 449.26            & 1699.09           & 463.61             & 1693.21            & 462.11            & 1716.94           & 461.75            & 1676.96           & \textbf{409.58}   \\
5-12-0.5-50-2                     & 7186.49      & 4100.27           & 6580.77           & 4105.01            & 6593.21            & \textbf{4099.23}  & 6510.86           & 4136.87           & 6581.53           & 4110.80           \\
5-12-0.5-50-4                     & 8116.56      & 6594.78           & 7798.70           & 7511.79            & 8951.90            & 7466.97           & 8638.58           & 6796.11           & 8046.86           & \textbf{6337.40}  \\
5-12-0.7-20-1                     & 975.55       & 134.31            & 739.67            & \textbf{127.94}    & 720.57             & 133.31            & 707.25            & 137.50            & 705.70            & 129.94            \\
5-12-0.7-20-2                     & 2822.93      & 1408.30           & 2573.14           & 1455.07            & 2525.03            & 1416.73           & 2582.26           & 1406.86           & 2597.28           & \textbf{1421.86}  \\
5-12-0.7-20-4                     & 2496.32      & 2376.68           & 2626.04           & 2549.59            & 2895.82            & 2224.94           & 2664.95           & 2158.18           & 2415.46           & \textbf{1535.03}  \\
5-12-0.7-50-1                     & 2271.82      & \textbf{238.48}   & 1746.99           & 240.86             & 1704.79            & 247.95            & 1736.11           & 246.16            & 1821.40           & 247.60            \\
5-12-0.7-50-2                     & 6692.71      & 3259.25           & 6104.51           & 3264.51            & 6165.50            & 3252.45           & 6162.88           & 3248.83           & 6081.45           & \textbf{3220.06}  \\
5-12-0.7-50-4                     & 6207.29      & 4818.25           & 6208.26           & 4765.20            & 6067.58            & 4901.68           & 6130.17           & 4729.12           & 5996.55           & \textbf{4553.07}  \\
\midrule
Avg.                              & 7470.74      & 6483.53           & 7320.69           & 7490.58            & 8367.35            & 7071.92           & 7880.41           & 7133.60           & 7943.18           & \textbf{5387.17}  \\ 
ARI$\uparrow$ & 10.72\%      & 22.51\%           & 12.51\%           & 10.48\%            & 0.00\%             & 15.48\%           & 5.82\%            & 14.74\%           & 5.07\%            & \textbf{35.62\%}          

\\ \bottomrule
\end{tabular}}
\end{adjustbox}
\end{table*}

This section evaluates the robust performance of EvoDR under different dynamic FAFSP operating conditions. A total of 480 instances across 24 operating conditions are constructed by varying the machine configuration $\mathrm{m} \in \{3 \times 6, 5 \times 12\}$, the flexibility factor $\varphi \in \{0.5, 0.7\}$, the number of newly arrived orders $\alpha \in \{20, 50\}$, and the load factor $\mu \in \{1, 2, 4\}$. Higher values of $\mu$ correspond to more frequent order arrivals. All 24 scenarios are evaluated through independent testing to assess the algorithm's performance under extreme conditions involving substantial shifts in data characteristics.

In addition to EDD, we evaluate eight hybrid heuristics formed by combining four job selection rules (FIFO, MOPNR, LWKR, and MWKR) with two machine assignment rules (SPT and EET)~\cite{lei2023large}. These nine configurations serve as the baseline methods against which EvoDR is benchmarked. For each operating condition, EvoDR is independently designed on the corresponding training instances and evaluated on the test instances. Experimental settings follow the descriptions in Sections~\ref{Exp:implementation} and~\ref{Exp:Hyperparametric}, and the detailed results are summarized in Table~\ref{tab: robust_experiment}.

Across the vast majority of the 24 conditions, EvoDR achieves the best or near-best performance, demonstrating its effectiveness in dynamic scheduling. Compared to the commonly used industrial rule EDD, EvoDR offers significant improvements. For instance, under condition (3-6-0.5-20-1), EvoDR achieves a tardiness of 1117.80 compared to 1710.45 for EDD, representing a 34.65\% reduction. Under a larger-scale instance (5-12-0.5-50-1), EvoDR reduces the tardiness from 2353.79 (EDD) to 409.58, achieving an improvement of 82.60\%. On average across all 24 scenarios, EvoDR yields the lowest average tardiness of 5387.17 and a substantial ARI of 35.62\%, significantly outperforming all baseline combinations.

Traditional heuristics show high sensitivity to operating conditions. For instance, FIFO+SPT performs competitively under low loads (e.g., 3-6-0.7-50-1), whereas MOPNR+EET degrades severely under high loads (e.g., 38635.80 under 3-6-0.5-50-4). In contrast, EvoDR maintains strong and stable performance across diverse environments, reducing tardiness by 16.91\% to 35.62\% over hybrid rules (e.g., averaging 5387.17 vs. FIFO+SPT's 6483.53 and MOPNR+EET's 8367.35). These results confirm that unlike sensitive handcrafted heuristics, EvoDR possesses strong generalizability and robustness against dynamic disturbances.


\section{Conclusion}
\label{conclusion}
This study investigates the automatic design of PDRs for the FAFSP with dynamic order arrivals under dual kitting constraints arising from assembly and multi-product delivery. An LLM assisted dynamic PDRs design framework named EvoDR is proposed to integrate algorithm design knowledge and scheduling analysis expertise for generating rules that adapt to evolving dynamic scheduling features. 
The effectiveness of the proposed method is validated through extensive experiments, including parameter sensitivity analysis and ablation studies. Results show that the fully integrated EvoDR outperforms all its ablated variants, demonstrating the effectiveness of the proposed modules.  
In comparative experiments on 20 instances from real production scenarios, EvoDR reduces average tardiness by 3.67-12.95\% compared to GP, GEP, and state-of-the-art LLM-based AHD methods, and by 16.81\% compared with the EDD baseline. 
Furthermore, robustness experiments conducted on 480 instances covering 24 combinations of resource load levels, disturbance intensities, and order conditions demonstrate that EvoDR outperforms the second best competitor by up to 13.11\%, indicating strong generalization and robustness under diverse conditions.

While EvoDR establishes a novel paradigm for dynamic PDR design, it currently relies on manual prompts and offline training. Future work will focus on reducing prompt dependency and extending the framework to handle complex constraints and concurrent disturbances such as secondary resources and transport limits. Moreover, while evolutionary search relies on feedback-driven diversity to maintain exploratory breadth, semantic guidance provides domain-specific directional cues via LLM reflection and feature profiling. Characterizing the individual and synergistic contributions of these two components to disentangle population exploration from heuristic semantic alignment remains a critical yet open theoretical question. Beyond manufacturing, EvoDR can be readily adapted to other problem domains by redefining state features and action spaces. Finally, exploring multi-stage integrated rule design and human-AI collaboration will further advance the vision of human-centered intelligent scheduling.

\bibliography{IEEEabrv,Manu}
\bibliographystyle{IEEEtran}

\clearpage
\clearpage
\appendix
\subsection{Related Works}
\label{sec:related_worlks}

\subsubsection{Flexible Assembly Flow Shop Scheduling}
The AFSP is a representative and typical combinatorial optimization problem in manufacturing systems.
According to \cite{framinan2019deterministic}, AFSP is defined as $\alpha 1 \rightarrow \alpha 2|\beta|\gamma$, where $\alpha 1$ and $\alpha 2$ represent the machine layouts for the processing and assembly stages, $\beta$ denotes special constraints, and $\gamma$ specifies the optimization objectives. While FAFSP is a multi-stage extension of AFSP, involving configurations like $DPm\rightarrow Fm$, $DPm\rightarrow HFm$, and $DFm\rightarrow 1$, where $Fm$ and $HFm$ refer to the flow shop and flexible/hybrid flow shop layouts in the assembly stage, and $DFm$ refers to a flow shop layout in the processing stage. Research on the $DPm\rightarrow Fm$ problem, particularly $DPm \rightarrow F2$, has produced several results~\cite{zhang2021maintenance, fernandez2022assembly}. However, $DPm\rightarrow HFm$, a more complex extension, still sees limited research~\cite{zeng2025constraint,framinan2019deterministic}.
The FAFSP addressed in this study is an extension of the $DPm\rightarrow HFm$ two-stage problem with the objective of minimizing tardiness. 
It incorporates the consolidated delivery constraint for multi-product orders in the final assembly stage, introducing significant challenges in temporal coordination, resource allocation, and process integration.

\subsubsection{Dynamic Production Scheduling based on PDRs}
Dynamic production scheduling requires rapid response to disturbances and timely schedule adjustments to prevent objective degradation. PDRs are widely favored in industry due to their high execution efficiency and interpretability.
The mainstream approaches for applying PDRs can be broadly categorized into PDR generation and PDR selection~\cite{branke2015automated}.
Generation focuses on the direct design of PDRs tailored to specific disturbance characteristics\cite{chen2025optimizing}. 
For instance, in FJSP and its variants, techniques such as GP and GEP are commonly employed to evolve rules for key decisions including job sequencing, machine allocation, and route planning~\cite{zhang2022multitask, zhou2020automatic, zhang2020evolving}. 
Furthermore, reinforcement learning-based PDRs selection is widely used by dynamically adapting to changes in the environment\cite{zhao2025deep}.
The scheduling problem is often formulated by such methods as a MDP, where actions are determined indirectly through the dynamic selection of PDRs. 
Subsequently, deep reinforcement learning is employed to train an agent on a predefined set of rules, thereby enabling the dynamic selection of rules~\cite{lei2023large, wang2022independent} or search strategies~\cite{cai2023novel} to address various scheduling problems. While this approach facilitates rule adjustment in response to state changes, its effectiveness is still highly dependent on the initial performance of the pre-selected PDR set.

\subsubsection{LLM-based AHD}
LLMs have introduced a novel paradigm for AHD, enabling the rapid construction of executable PDRs through LLMs. Notable examples include EoH~\cite{EOH} and MEoH~\cite{meoh}, which incorporate multiple prompting strategies and dominance-based selection to enhance the diversity and efficiency of heuristic generation. FunSearch established an early template by integrating LLMs within an island-based evolutionary framework for code optimization~\cite{funsearch}. Moreover, alternative search paradigms such as Monte Carlo tree search~\cite{mcts_ahd, qiu2026e2oc} and neighborhood search~\cite{xie2025llm} have also been adapted to structure the exploration process and improve sample efficiency in heuristic synthesis. Collectively, these approaches reduce reliance on manually predefined symbolic sets and operate in semantically richer spaces, enabling more flexible heuristic design.

Beyond generation and search, reflective mechanisms have recently been integrated into LLM-based AHD to enable dynamic self-correction. Representative works such as ReEvo~\cite{reevo} embed reflection into evolutionary search, where LLMs compare algorithm variants and extract insights to guide subsequent generations. Similarly, outer-layer evolutionary prompts~\cite{ye2025large} have been introduced to maintain population diversity. These approaches typically follow a generate-reflect-revise loop~\cite{shinn2023reflexion}, focusing on identifying errors or deficiencies in a single generated output and repairing it within the same iteration. While effective for local improvement, this corrective mechanism operates at the individual level and does not explicitly leverage population-wide knowledge or domain-specific feature interactions to guide the search direction.



\subsection{Formulation of Dynamic PDRs Optimization}
\label{Optimization_formulation_appendix}
Let $\mathcal{X}$ denote the set of training instances, each $\mathrm{x} \in \mathcal{X}$ representing a specific manufacturing configuration with distinct resource structures, disturbance distributions, and product constraints. Given a PDR $\pi$, the tardiness on instance $\mathrm{x}$ is defined as

\begin{equation}
\label{eq_pdrobj}
F(\mathrm{x} \mid \pi) = f(\mathrm{x} \mid \mathbf{S}(\mathrm{x} \mid \pi)), \quad \forall \mathrm{x} \in \mathcal{X},
\end{equation}

where $\mathbf{S}$ is the dynamic scheduler that generates a schedule under rule $\pi$, and $f$ computes the resulting order tardiness. The average tardiness over the instance set is $F(\mathcal{X}\mid\pi)=\frac{1}{|\mathcal{X}|}\sum_{\mathrm{x}\in\mathcal{X}}F(\mathrm{x}\mid\pi)$, and the optimization objective is

\begin{equation}
\label{eq_evo_obj}
\pi^* = \underset {\pi \in \mathbf{\Pi}} {\text{arg min}} F(\mathcal{X} \mid \pi),
\end{equation}

subject to a total computational budget $T$. Here, $\mathbf{\Pi}$ denotes the space of all feasible PDRs.

\subsection{Encoding}
\label{sec:encoding}
Each individual is composed of a different semantic textual description of a heuristic rule containing the corresponding description, code, and performance information, shown as A1 in Fig.~\ref{figure_iteration_process}. 
The code implements a specific PDR, which operates at step $t$ to sort job-machine pairs and filter the set of candidate actions for execution in the subsequent step.

Specifically, each rule individual generated by LLM-A consists of four components: 
1) Textual description, which summarizes the rule's fundamental idea. 
2) Code block, which provides the executable implementation. The code strictly complies with specified function names and input–output formats to ensure portability and compatibility across different scheduling environments. 
3) Objective fitness score $\mathrm{c}^s$, the numerical performance of the objective computed by applying the rule to a simulator with multiple instances. 
4) Subjective evaluation and suggestion $\mathrm{c}^e$, which are generated by LLM-S to analyze the design ideas, code and numerical performance of the PDR.

\subsection{Elite Knowledge Guided Rules Generation}
\label{sec:appendix_algorithm}
\begin{algorithm}[!t]
\caption{Elite knowledge guided rules generation}
\begin{algorithmic}[1]
    \STATE \textbf{Input:}
    Instance set $\mathcal{X}$, elite rule $\pi^0$, population size $\mathcal{P}$, LLM's temperature $u$ and range $(U_{low},U_{up})$.
    \STATE \textbf{Output:}
    Initial PDR population $P^0$ and temperature $u_0$.
    \STATE Initialize the simulator $\mathbf{E}(\cdot)$, $\mathbf{G}^{(A)}(\cdot)$ and $\mathbf{G}^{(S)}(\cdot)$;
    \STATE Get the prompt elements $(\mathrm{C}^A/\mathrm{C}^S,\mathrm{c}^f,\mathrm{c}^c,\mathrm{c}^s,\mathrm{c}^e)$;
    \STATE $F(\mathcal{X} \mid \pi^0) \leftarrow \mathbf{E}(\mathcal{X} \mid \pi^0)$;
    \STATE $\mathrm{c}^f \leftarrow$ calculate the dynamic features;
    \STATE $\mathrm{c}^c,\mathrm{c}^s,\mathrm{p},F_{best}\leftarrow \pi^0,F(\mathcal{X} \mid \pi^0),1,V$;
    \STATE $p^S \leftarrow (\mathrm{C}^S,\mathrm{c}^f,\mathrm{c}^c,\mathrm{c}^s)$.
    \STATE $\mathrm{c}^e \leftarrow \mathbf{G}^{(S)}(\cdot \mid p^S)$;
    \WHILE{$\mathrm{p} \le \mathcal{P}$}
    \STATE $u \leftarrow U_{low} +\mathrm{p} \times (U_{up}-U_{low})/{\mathcal{P}}$;
    \STATE $p^A \leftarrow (\mathrm{C}^A,\mathrm{c}^f,\mathrm{c}^c,\mathrm{c}^s,\mathrm{c}^e)$;
    \STATE $\pi' \leftarrow \mathbf{G}^{(A)}(\cdot \mid (p^A,u))$;
    \IF{$\pi'$ is valid}
    \STATE $F(\mathcal{X} \mid \pi') \leftarrow \mathbf{E}(\mathcal{X} \mid \pi')$;
    \STATE $P^0_\mathrm{p},\mathrm{p} \leftarrow (\pi',F(\mathcal{X} \mid \pi')),\mathrm{p}+1 $;
    \IF{$F(\mathcal{X} \mid \pi') \leq F_{best}$}
    \STATE $u_0,\pi^*, F_{best} \leftarrow u,\pi',F(\mathcal{X} \mid \pi')$;
    \ENDIF
    \ENDIF
    \ENDWHILE
\end{algorithmic}
\label{alg:Elite_rules_generation}
\end{algorithm}

\subsection{Instance Generation Details}
In the dynamic FAFSP, the machines and jobs in the processing stage can be represented as $m_{1} \in M_p$ and $j_1 \in J_p$, while the assembly lines and jobs are represented as $m_{2} \in M_a$ and $j_2 \in J_a$. 
The product structure is extracted from the actual production data, and the machine qualification for each job is determined by a flexibility factor $\varphi$. 
According to manufacturer statistics, the probability distribution of the number of products $q$ of each order is given by $P(X=q)=(1/2)^q$. 
The processing rate of the assembly lines, $V_{a}$, follows a normal distribution $V_{a} \sim \mathcal{N}(8, 2)$, the average processing rate $V_{p}$ in the processing stage is set to $0.8 V_{a}$. 
Order due dates follow a uniform distribution $[L(1-T-R/2), \; L(1-T+R/2)]$, in accordance with~\cite{basir2018bi}, where $L$ denotes the lower bound of the makespan computed as in Eq.~\ref{eq42}. Then $T$ and $R$ are hyperparameters controlling the range of $DT_i$, usually set between 0 and 1. 
\begin{equation}
\label{eq42}
\begin{aligned}
L = \max \biggl\{ 
 \max_{m_1\in M_p} \sum_{j_1}^{J_p} PT_{j_1,m_1} + \min_{m_2\in M_a} \sum_{j_2}^{J_a} PT_{j_2,m_2}, \\
 \max_{m_2\in M_a} \sum_{j_2}^{J_a} PT_{j_2,m_2} 
\biggr\}
\end{aligned}
\end{equation}

Initially, a subset of orders already exists, while subsequent arrivals follow a Poisson process. 
In decision step $t$, the processing and assembly jobs are represented as $J_p^{(t)}$ and  $J_a^{(t)}$, and the total number of orders that have arrived is represented as $I^{(t)}$.
Specifically, the inter-arrival time $\Delta \mathrm{g}$ between consecutive orders follows an exponential distribution with mean $\lambda$, which is dynamically calculated as the estimated completion time under load for the average of orders that have arrived in step $t-1$, as defined in Eq.~\ref{eq43} and~\ref{eq44}.
The number of dynamically arriving orders is $\alpha \in \{20, 50\}$, and the load factor is $\mu$, and the order arrival times are defined by Eq.~\ref{eq45}. To evaluate the performance of EvoDR under different order arrival frequencies and production loads, instances with different operating conditions are generated by adjusting $(m_1, m_2, \varphi, \alpha, \mu)$, and various experiments are designed for a comprehensive evaluation.

\begin{equation}
\label{eq43}
\begin{aligned}
ET^{(t)} = \sum_{i=1,\ldots,I^{(t)}} \biggl(  \max_{m_1\in M_p} \Big( \sum_{j=1,\ldots,I_p^{(t)}} PT_{j,m_1} \times (B_{i,j} - 1) \Big) \\ + \min_{m_2 \in M_a} \Big( \sum_{j=1,\ldots,I_a^{(t)}} PT_{j,m_2} \times (B_{i,j} - 1) \Big) \biggr)
\end{aligned}
\end{equation}
\begin{equation}
\label{eq44}
\lambda = ET^{(t-1)}/(I^{(t)}\times \mu )
\end{equation}
\begin{equation}
\label{eq45}
AT_i = AT_{i-1}  + \Delta \mathrm{g}
\end{equation}

\subsection{Temperature Sensitivity Analysis}
Temperature range $(U_{low},U_{up})$ settings are used to enhance the diversity of the initial PDRs and to determine the evolution temperature in Algorithm~\ref{alg:Elite_rules_generation}. Following the recommendations from related studies and from model providers, five temperature ranges are examined, including $(0, 2.0)$, $(0, 1.6)$, $(0, 1)$, $(0.3, 1.5)$, and $(0.5, 1.25)$. The experimental setup is the same as described above, and the results are reported in Table~\ref{tab: Us}.

The results show that EvoDR's performance remains stable across temperature configurations, with $T_{all}$ varying within a narrow band of 1139--1197 across all five ranges. The adaptive temperature search mechanism selects effective $u$ values between 0.33 and 0.94 depending on the range setting. The range $(0.3, 1.5)$ achieves the best overall performance with $T_{all}=1139.55$, followed by $(0.5, 1.25)$ with $T_{all}=1142.84$. Expanding the upper bound to 2.0 yields a $T_{all}$ of 1197.22, indicating that higher temperatures yield no additional benefit, as the adaptive mechanism avoids the less effective high-temperature region. Based on these results, $(0.3, 1.5)$ is adopted as the default configuration in subsequent experiments. All 24 cross-scenario experiments use this same default range, without any per-scenario temperature tuning.

\begin{table}[h]
\caption{Comparison of different temperature ranges.}\label{tab: Us}
\centering
\resizebox{\columnwidth}{!}{
\begin{tabular}{cccccc}
\toprule
$(U_{low},U_{up})$ & $T_{train}$           & $T_{test}$             & $T_{all}$             & $u$       & Acc              \\
\midrule
(0, 2)                & 1041.15±90.07         & 1353.29±50.42          & 1197.22±62.35         & 0.67±0.23 & 92.26\%          \\
(0, 1.6)              & 1004.23±29.63         & 1312.88±67.22          & 1158.56±19.18         & 0.64±0.32 & 94.09\%          \\
(0, 1)                & 1025.20±66.64         & 1356.91±26.29          & 1191.06±45.40         & 0.33±0.42 & \textbf{95.92\%} \\
(0.3, 1.5)             & 1013.00±78.45         & \textbf{1266.10±69.66} & \textbf{1139.55±7.40} & 0.94±0.55 & 80.06\%          \\
(0.5, 1.25)            & \textbf{978.03±20.98} & 1307.64±61.56          & 1142.84±35.29         & 0.90±0.17 & 90.54\%         
       \\
\bottomrule
\end{tabular}}
\end{table}

\subsection{Cost Analysis}
\textbf{Cost analysis:} As shown in Table~\ref{tab: cost}, EvoDR with full LLM-S achieves the best performance (All: 1185.1, Acc: 89.3\%) at a total cost of \$0.189, comprising \$0.123 for LLM-A and \$0.066 for LLM-S. Consuming a mere 0.16M tokens per run, the LLM-S component powers iterative evaluation to drive superior results. Compared with EoH (1297.4) and ReEvo (1242.9), EvoDR demonstrates that the modest additional cost of LLM-S yields substantial performance gains, establishing the best overall cost-effectiveness. For deployment, since sampling volume is the primary cost driver, refining prompt quality to accelerate convergence offers the most practical path to further savings. Furthermore, it is important to distinguish between the one-time offline design cost (roughly \$0.189 per run for LLM API calls and simulation) and the online dispatching cost. The online phase requires only a single priority ranking over the candidate job--machine pairs per decision, completely bypassing LLM calls and executing in sub-milliseconds.

\begin{table}[h]
\caption{Comparison of costs for different methods.}\label{tab: cost}
\centering
\resizebox{1\columnwidth}{!}{
\begin{threeparttable} 
\begin{tabular}{lcccccc}
\toprule
Method & $T_{train}$    & $T_{test}$      & $T_{all}$       & Token                & Acc             & Cost                 \\
\midrule
EoH    & 1161.2         & 1433.6          & 1297.4          & 0.32M                & 87.3\%          & 0.131                \\
ReEvo  & 1053.2         & 1432.5          & 1242.9          & 1.36M                & 70.4\%          & 0.558                \\
EvoDR  & \textbf{987.8} & \textbf{1382.4} & \textbf{1185.1} & \textbf{0.30M/0.16M} & \textbf{89.3\%} & \textbf{0.123/0.066}
\\
\bottomrule
\end{tabular}   
\begin{tablenotes}    
    \footnotesize            
   \item  All methods sample 100 rules, and the experiment is repeated three times. EvoDR additionally requires the evaluation of both individuals and populations. The average number of samplings for LLM-S is 115, with Token and Cost denoted as -A / -S.
\end{tablenotes}     
\end{threeparttable}}
\end{table}

\subsection{Further Analysis of the Ablation Experiment}
\begin{figure}[h]
\centering
\makebox[\columnwidth][c]{\includegraphics[width=1\columnwidth]{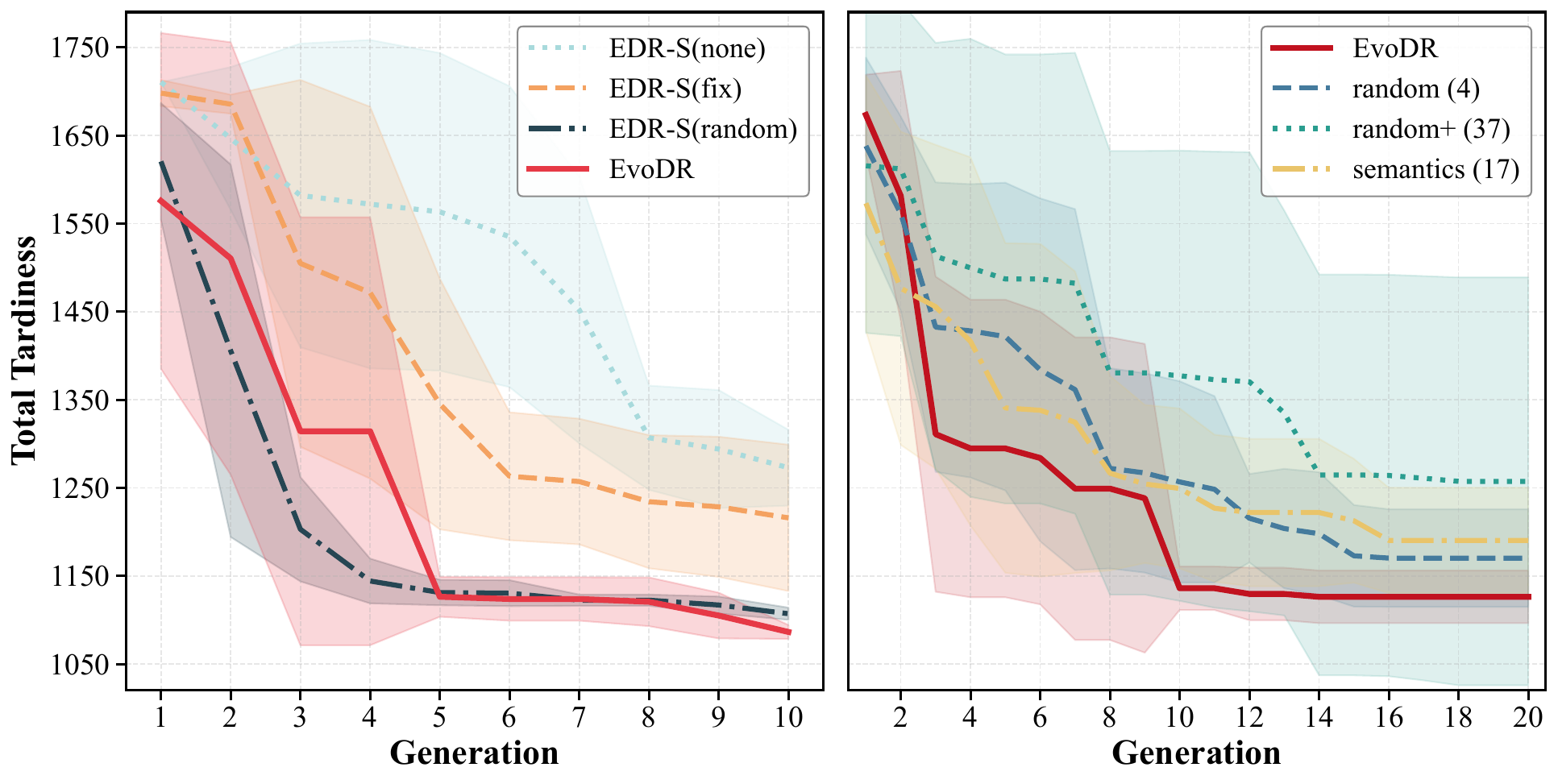}}
\caption{Convergence analysis for 4 (left) and 37 (right) suggestions.}
\label{figure_llmsconvergence}
\end{figure}

To understand this effect, Fig.~\ref{figure_llmsconvergence} further examines the convergence of the random-feedback variant under two suggestion-pool sizes: a small pool of 4 suggestions (left) and an expanded pool of 37 suggestions (right, evaluated on the training set). While the small pool can yield competitive results, the expanded pool reveals that random feedback becomes markedly less stable and its performance degrades as the number of suggestions grows. This indicates that the benefit of random feedback is not stable, and that the scheduling-specific semantic feedback of LLM-S provides an additional, directed value.

Furthermore, feature sensitivity analysis shows that under dual kitting constraints, EvoDR is highly sensitive to structural features. Eliminating arc features (1630.28) and order features (1579.27) leads to disastrous performance drops, whereas omitting machine features (1140.60) maintains competitive results. This stark contrast suggests that capturing topological and job-level dynamics is far more critical than focusing on processing units alone.

\subsection{Further Analysis of the Comparative Experiments}
\begin{figure}[h]
\centering
\makebox[\columnwidth][c]{\includegraphics[width=1\columnwidth]{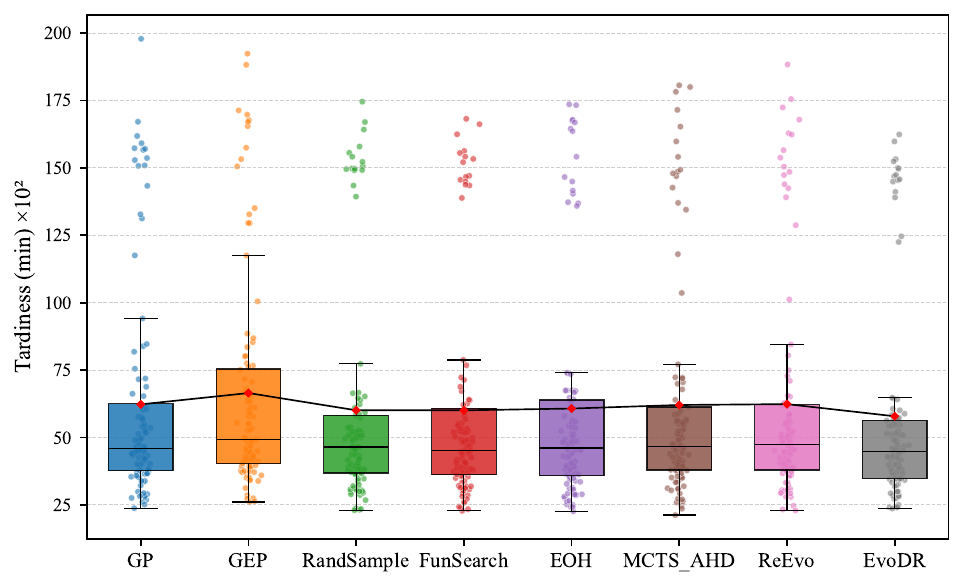}}
\caption{Box plot of tardiness performance in different AHD methods.}
\label{figure_boxplot_AHDs}
\end{figure}

The performance of all AHD methods are shown in Fig.~\ref{figure_boxplot_AHDs}. The red markers represent the average performance of the different methods, and the black lines indicate the median.

\subsection{Further Analysis of the Robustness Experiment}
\begin{figure*}[t]
\centering
\makebox[\textwidth][c]{\includegraphics[width=0.8\textwidth]{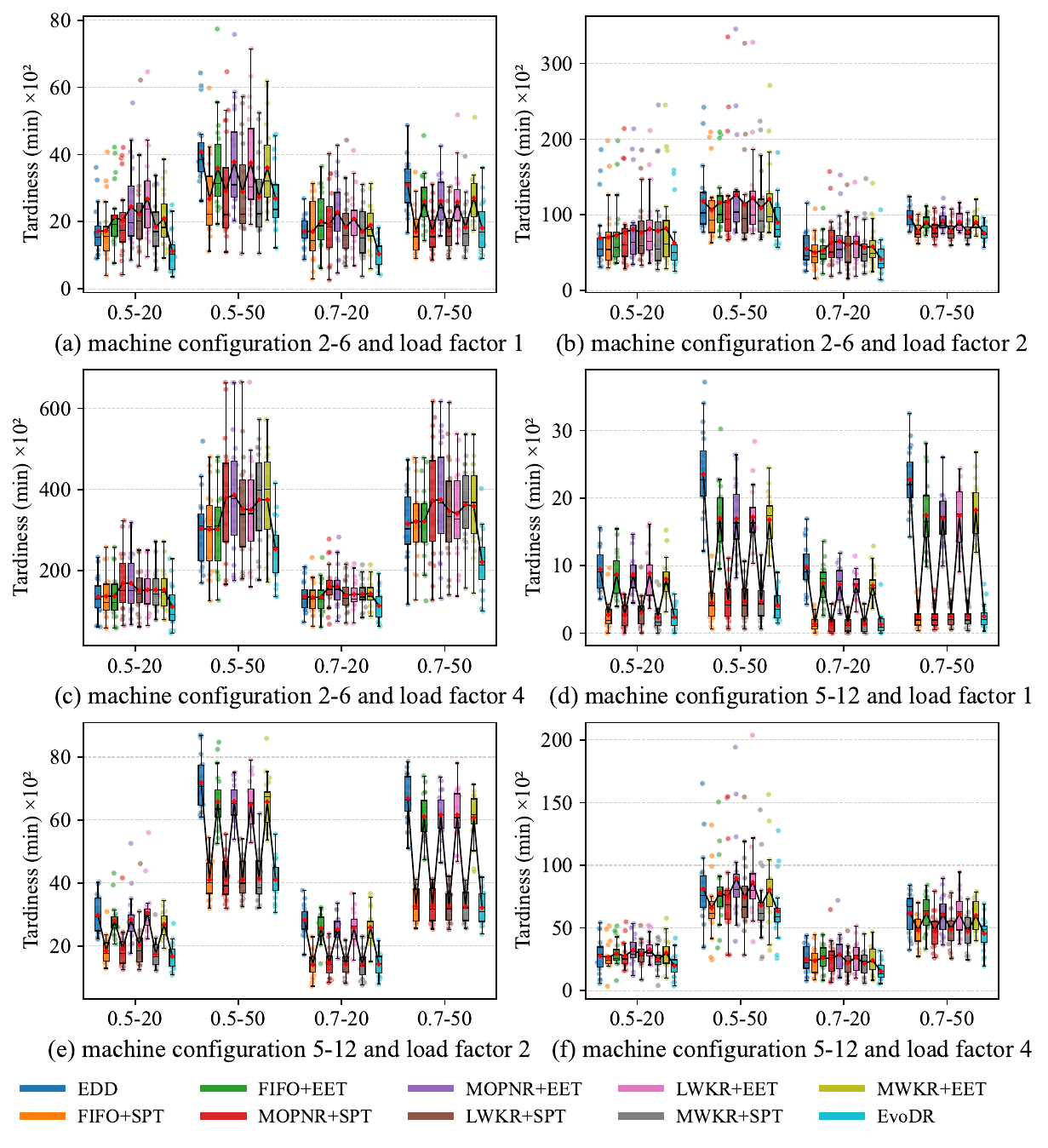}} 
\caption{Box plot of tardiness performance under different conditions.}
\label{figure_rb_results}
\end{figure*}

The boxplots of the performance of all the methods under different conditions are shown in Fig.~\ref{figure_rb_results}. Under identical resource configurations, higher load factors $\mu$ lead to more frequent order arrivals, intensifying resource contention and resulting in higher tardiness. For instance, in experiments with (3-6-0.5-20-$\mu$) where $\mu$ increases from 1 to 2 and 4, the average tardiness of EvoDR increases from 1117.80 to 6265.49 and 11079.05, respectively. Similar trends are observed in other scenarios, indicating that higher order arrival frequency imposes greater demands on the algorithm’s responsiveness. Crucially, EvoDR demonstrates exceptional robustness under high-load conditions. For example, under (3-6-0.7-50-4), EvoDR achieves a tardiness of 22108.98, reducing the total tardiness by 9952.04 compared to the best-performing traditional heuristic in the same scenario (LWKR+EET: 34045.23), which corresponds to a significant performance leap.

\end{document}